\PassOptionsToPackage{table}{xcolor}
\documentclass{article}

\usepackage[preprint]{neurips_2026}

\usepackage[utf8]{inputenc}
\usepackage[T1]{fontenc}
\usepackage{mathtools}
\usepackage{amssymb}
\usepackage{bm}
\usepackage{xcolor}
\usepackage{graphicx}
\usepackage{booktabs}
\usepackage{tabularx}
\usepackage{multirow}
\usepackage{makecell}
\usepackage{array}
\usepackage{float}
\usepackage{wrapfig}
\usepackage{caption}
\usepackage{subcaption}
\usepackage{microtype}
\usepackage{url}
\usepackage{hyperref}
\usepackage[nameinlink,capitalise]{cleveref}
\usepackage{tikz}
\usetikzlibrary{arrows.meta,positioning,fit,backgrounds}
\usepackage{pgfplots}
\pgfplotsset{compat=1.18}

\setcitestyle{authoryear,open={(},close={)}}
\hypersetup{colorlinks=true,allcolors=blue}

\newcommand{\algacro}{\textsc{ScreenSearch}}
\newcommand{\ind}{\mathbb{I}}
\newcommand{\pospart}[1]{\left[#1\right]_{+}}
\newcolumntype{Y}{>{\centering\arraybackslash}X}

\title{\algacro{}: Uncertainty-Aware OS Exploration}

\author{
Michael Solodko$^{*\dagger}$ \quad Justin Wagle$^{*\dagger}$\\
Microsoft$^{\dagger}$\\
\small $^{*}$Equal contribution.
}

\begin{document}
\maketitle

\begin{abstract}
Desktop GUI agents operate under partial observability: nearly identical screens can hide different workflow states, and the same interaction can therefore have different outcomes. We present \algacro{}, a distributed exploration system that builds a shared deduplicated screen graph and treats matched-action outcome dispersion as a structural ambiguity signal. A one-step PUCT graph-bandit combines state and edge novelty with movement toward lower-dispersion states; it executes one action per selection and performs no simulated rollouts or multi-step backup. Across 11 applications, \algacro{} records 1,007,406 screenshots, 39,902 unique observations, and 31,146 merged graph states. In a matched evaluation with 50 replay starts in each of Notepad, File Explorer, and LibreOffice Calc, PUCT raises macro-average new-state AUC from 4.820 to 50.487 over greedy selection using the same prior. Adding the ambiguity term to novelty raises state AUC by 8.9\%, while trading off against edge coverage, so we do not claim uniform dominance. Controlled hidden-state, timing-noise, and reversible-probe tests clarify what the ambiguity proxy detects and when lower dispersion constitutes useful disclosure. Finally, an OCR audit shows that the representation need not be tied exclusively to accessibility metadata, and a downstream policy study shows that counterfactual screen supervision improves state sensitivity and offline recovery. These results support uncertainty-aware state exploration while making its causal and deployment limits explicit.
\end{abstract}

\section{Introduction}

Computer-use environments are large, heterogeneous, and partially observable. A single desktop observation rarely determines the underlying workflow state: near-identical screens can reflect different permissions, clipboard contents, focus, authentication context, pending asynchronous events, or hidden application modes. The challenge is therefore not only to choose a plausible next UI action, but also to decide when the available observation is sufficient for commitment and when another interaction should gather evidence. A locally reasonable action can otherwise push an agent into a brittle part of the workflow before the relevant latent cases have been distinguished.

Most computer-use agents are evaluated through end-to-end instruction following and operate stepwise or with a short receding horizon. That paradigm is effective for task completion, but offers limited support for explicitly mapping an unfamiliar application or uncovering the hidden structure behind repeated observations. We instead study open-ended computer/OS exploration, where the product of interaction is a reusable graph of screens, actions, and observed outcomes. In this setting, observations are better treated as hypotheses about workflow state than as complete state descriptions.

Two exploration signals are complementary. \emph{Novelty} rewards discovering new deduplicated states and labeled transitions, thereby expanding the known frontier. \emph{Ambiguity} measures whether matched actions from similar observations lead to inconsistent successors. Novelty alone can over-reward superficial visual change; ambiguity alone can remain within a small uncertain region or prefer an uninformative escape. Their useful balance depends on the proposal mechanism, the quality of state identity, and the cost of committing while outcomes remain unreliable.

We introduce \algacro{}, a distributed system that turns UI observations into location-aware structural and semantic features, retrieves candidate matches with a hybrid sparse--dense index, verifies near-duplicates, and shares a state graph across isolated VM workers. This global memory supports revisit detection, loop avoidance, action-level statistics, and empirical ambiguity estimation. Over the graph, a one-step PUCT selector balances plausible actions, frontier expansion, and movement toward lower-dispersion states. We call it a \emph{PUCT graph-bandit}: it borrows the PUCT selection rule, but executes one environment action per selection and performs neither internal simulations nor multi-step return backup.

The same interaction traces support learned action proposals and downstream screen-conditioned policies. We retain structured screen--action windows, remove loop- and no-op-dominated traces, and construct same-task/different-screen counterfactual pairs. This connects exploration to policy learning without requiring the explorer itself to become a trajectory-level planner: learned proposals guide which executable actions deserve attention, while downstream models test whether the collected data teaches actions that depend on the current screen.

We evaluate the full system at two complementary scales. A 5,524-VM-hour corpus across 11 applications measures state discovery and operational behavior. A matched 150-start suite across Notepad, File Explorer, and LibreOffice Calc isolates selector, objective, proposal, and aggregation effects. Controlled clipboard, timing, and reversible-action studies distinguish hidden-condition dispersion from stochastic execution and test whether a candidate probe improves a later commitment. Finally, OCR and downstream-policy studies examine whether the representation and collected traces transfer beyond the exact exploration backend.

Our contributions are:
\begin{itemize}
    \item a scalable distributed screen graph built from location-aware UI features, hybrid sparse--dense retrieval, bounded near-duplicate verification, and deterministic action identities;
    \item a matched-action outcome-dispersion proxy and one-step PUCT objective for uncertainty-aware exploration, accompanied by causal controls and an operational probe criterion;
    \item large-scale exploration across 11 desktop applications plus matched cross-application selector, objective, prior, state-aggregation, and sensitivity analyses; and
    \item a trace-processing bridge from exploration to learned proposals and downstream policies, with evidence that OCR can supply an alternative representation and counterfactual screen supervision improves offline grounding and recovery.
\end{itemize}

\section{Related work}

\paragraph{Interactive web and OS agents.}
Computer-use benchmarks range from programmatic web interfaces to realistic websites and full desktop environments \citep{liu2018miniwob,deng2023mind2web,zhou2023webarena,xie2024osworld}. They expose brittleness under long horizons, UI variation, and real execution constraints. We share the goal of robust interaction, but focus on the complementary systems problem of discovering reusable state structure when observations are aliased, rather than optimizing a single task-completion episode.

\paragraph{Perceptual aliasing and active sensing.}
Perceptual aliasing describes observations that fail to distinguish states requiring different behavior \citep{chrisman1992aliasing}. Active sensing and active perception choose actions partly for the evidence they reveal \citep{veiga2023activesensing,bajcsy2018activeperception}. These methods often optimize entropy or mutual-information objectives under explicit belief-state or world-model assumptions. \algacro{} adopts their information-seeking motivation without claiming a Bayesian posterior: its ambiguity score is a scalable empirical reliability proxy over matched action outcomes.

\paragraph{Planning and exploration.}
POMCP and AlphaZero-style search show how visit statistics and priors can guide decisions in large spaces \citep{silver2010pomcp,silver2017mastering,silver2018general}. Intrinsic-motivation methods reward prediction error, novelty, visitation, or learned uncertainty \citep{pathak2017curiosity,burda2018rnd,sekar2020plan2explore}. GUI environments add a practical difficulty: visually similar observations can hide different workflow states, so state novelty need not imply reliable action outcomes. Our selector uses the PUCT exploration bonus over a transposition graph, but updates it only with the observed immediate transition reward. It is therefore neither multi-step MCTS nor belief-state planning.

\paragraph{Stepwise agents and retrieval-augmented memory.}
Language agents commonly interleave short-horizon reasoning and tool actions \citep{yao2022react}; reflection and deliberate-search variants improve robustness by reconsidering candidate continuations \citep{ahn2022saycan,shinn2023reflexion,zhou2023lats}. Other systems retrieve relevant prior experience as an external memory \citep{kagaya2024rap}. \algacro{} uses retrieval for a different but related purpose: it identifies revisits and near-duplicates, pools matched-action evidence, and exposes shared graph statistics to many workers. The resulting memory is part of the exploration state rather than only prompt context.

\paragraph{Trajectory learning.}
Offline sequence models treat control as conditional generation, while diffusion policies and planners represent multimodal actions through denoising \citep{chen2021decisiontransformer,janner2022diffuser,chi2023diffusionpolicy}. We use \algacro{} traces both to learn proposal priors and to study screen-conditioned action prediction. In contrast to using diffusion as the explorer's search procedure, the reported exploration policy remains the one-step graph-bandit; the downstream diffusion evaluation is an offline, teacher-forced proxy rather than live task completion.

\section{Computer/OS state exploration}
\label{sec:formulation}

\begin{wrapfigure}{r}{0.40\textwidth}
    \centering
    \vspace{-0.8\baselineskip}
    \includegraphics[width=0.98\linewidth]{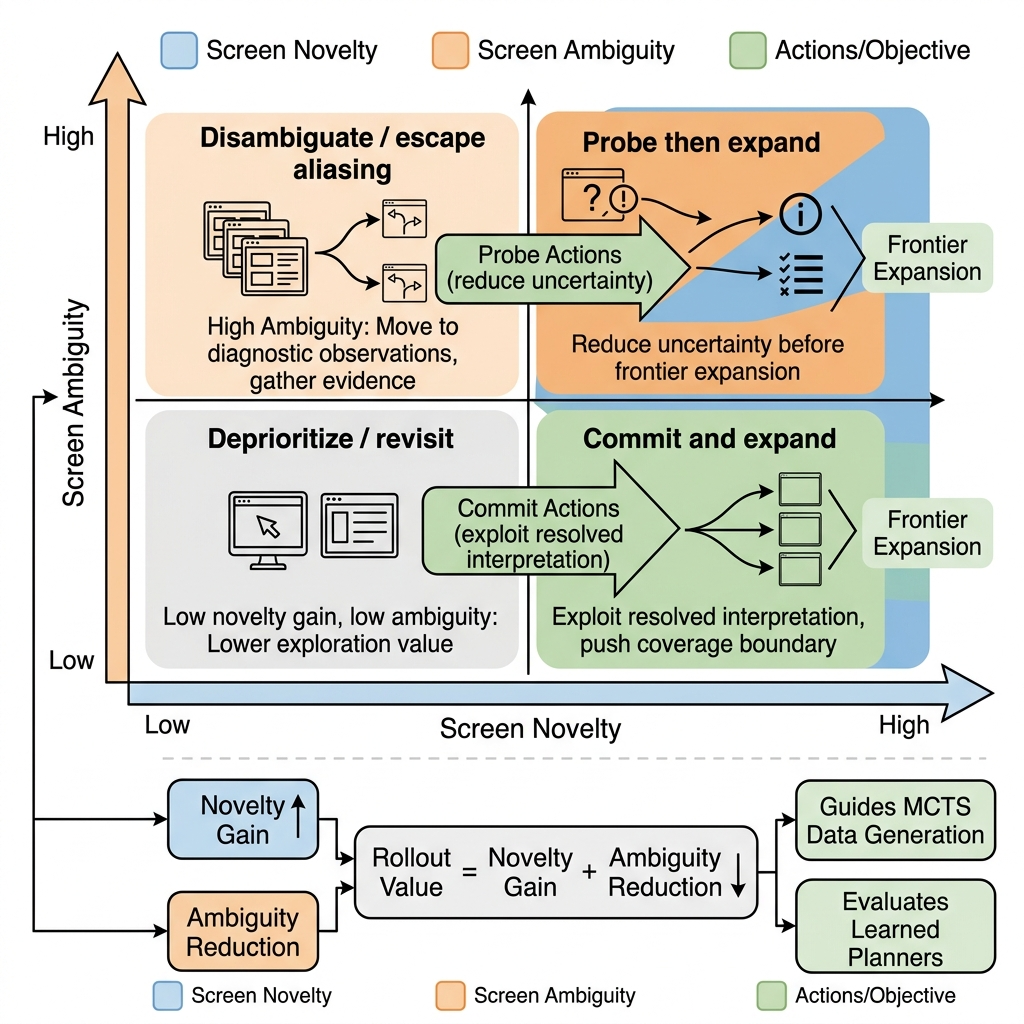}
    \vspace{-0.55\baselineskip}
    \caption{Complementary exploration signals: novelty expands coverage, while ambiguity identifies empirically unreliable matched-action outcomes.}
    \label{fig:ambiguity_vs_novelty}
    \vspace{-0.8\baselineskip}
\end{wrapfigure}

Let $G_t=(V_t,E_t)$ be the global deduplicated state graph before interaction $t$. A realized labeled edge is $(s,\sigma,s')$, where $s$ and $s'$ are merged states and $\sigma$ is a deterministic action signature. A transition is state-novel when $s'\notin V_t$ and edge-novel when $(s,\sigma,s')\notin E_t$.

\paragraph{Screen novelty and the exploration frontier.}
State and edge novelty capture different forms of progress. A new state expands the set of known observations, whereas a new labeled edge reveals another reachable outcome or another way to reach an existing state. Retaining both matters in GUIs: two actions may converge to the same dialog while exposing different interaction affordances, and a visually new screen may be reached through a transition already represented elsewhere in the graph. Novelty is always evaluated against the shared graph snapshot immediately preceding an action, so concurrent workers contribute to a common, continually changing frontier.

\paragraph{Action identity.}
The exact signature is
\begin{center}
\small\nolinkurl{template_id@version::target_signature::arg_signature}.
\end{center}
The target signature is \texttt{cell|role|label|application}: the target center is quantized on a $30\times30$ active-window grid, and role, label, and application strings are normalized. Arguments use sorted compact JSON. Clicks therefore retain target location and semantics; type actions additionally retain the exact text. Candidate IDs and raw pixels are excluded because they are unstable. This prevents graph counts and ambiguity pools from collapsing parameter-distinct interactions. Examples appear in \cref{app:reproducibility}.

\paragraph{Screen ambiguity.}
Let $\mathcal{O}(s)$ be the near-duplicate observation cluster assigned to state $s$, and let $\Sigma(s)$ be its observed signatures. For each signature, $P(s'\mid s,\sigma)$ is the empirical successor distribution and $n_{s,\sigma}$ its sample count. We define
\begin{align}
D(s)&=\sum_{\sigma\in\Sigma(s)}w_{s,\sigma}\,\bar H\!\left(P(s'\mid s,\sigma)\right),
&w_{s,\sigma}&=\frac{n_{s,\sigma}}{\sum_{\tilde\sigma}n_{s,\tilde\sigma}},\\
\rho(s)&=\frac{n_s}{n_s+\kappa},
&u(s)&=\rho(s)D(s)+(1-\rho(s))u_0,
\label{eq:ambiguity}
\end{align}
where $\bar H\in[0,1]$ is normalized entropy and $n_s=\sum_\sigma n_{s,\sigma}$. Reported runs use $\kappa=20$ and $u_0=0.5$. The score detects empirical outcome dispersion; it does not identify its cause from a single observation. Hidden workflow state, asynchronous timing, execution noise, and merge errors can all raise $u(s)$.

\paragraph{Joint exploration objective.}
Pure novelty can repeatedly chase cosmetic changes, while pure ambiguity reduction may remain local or simply leave an uncertain region. We therefore reward graph expansion and positive ambiguity reduction together. This is a structural information-seeking objective rather than an explicit expected-information-gain calculation: it is designed to remain computable from the transition log shared by the workers, without learning a latent dynamics model or maintaining a belief distribution over every application state.

\paragraph{Probe versus escape.}
We do not label an action a probe merely because it reaches a low-dispersion destination. Operationally, a probe must reveal information that improves a later latent-dependent commitment relative to a matched non-disclosing action. This criterion is evaluated retrospectively in the controlled study in \cref{sec:controlled}; it is not a hard-coded action class used by the selector.

\section{Representation, retrieval, and state identity}
\label{sec:retrieval}

\subsection{Screen representation}

Each screen observation $S$ is represented from its UI Automation (UIA) element set $\mathcal{E}(S)$. An element supplies a bounding box, control role, and optional text. We normalize the role and text, quantize the element center into a shared spatial grid, and create cell--control and cell--text atoms. These atoms retain local layout and lexical cues while tolerating small coordinate shifts. Display mode and text-size bin are stored as screen-scoped metadata, and concatenated normalized screen text is embedded with OpenAI \texttt{text-embedding-3-small}.

The resulting mixed signature has three roles. Sparse atoms support interpretable lexical retrieval and bounded overlap tests; the dense vector recovers semantically related screens whose surface tokens differ; and the metadata prevents implausible matches across configured display contexts. UIA is the backend used for the reported online runs, but indexing, graph construction, and ambiguity estimation consume the resulting feature sets rather than invoking UIA-specific logic. This separation makes it possible to substitute another observation encoder, as tested by the OCR audit in \cref{app:representation_audit}.

\begin{figure}[t]
\centering
\begin{subfigure}[t]{0.33\linewidth}
    \centering
    \includegraphics[width=\linewidth]{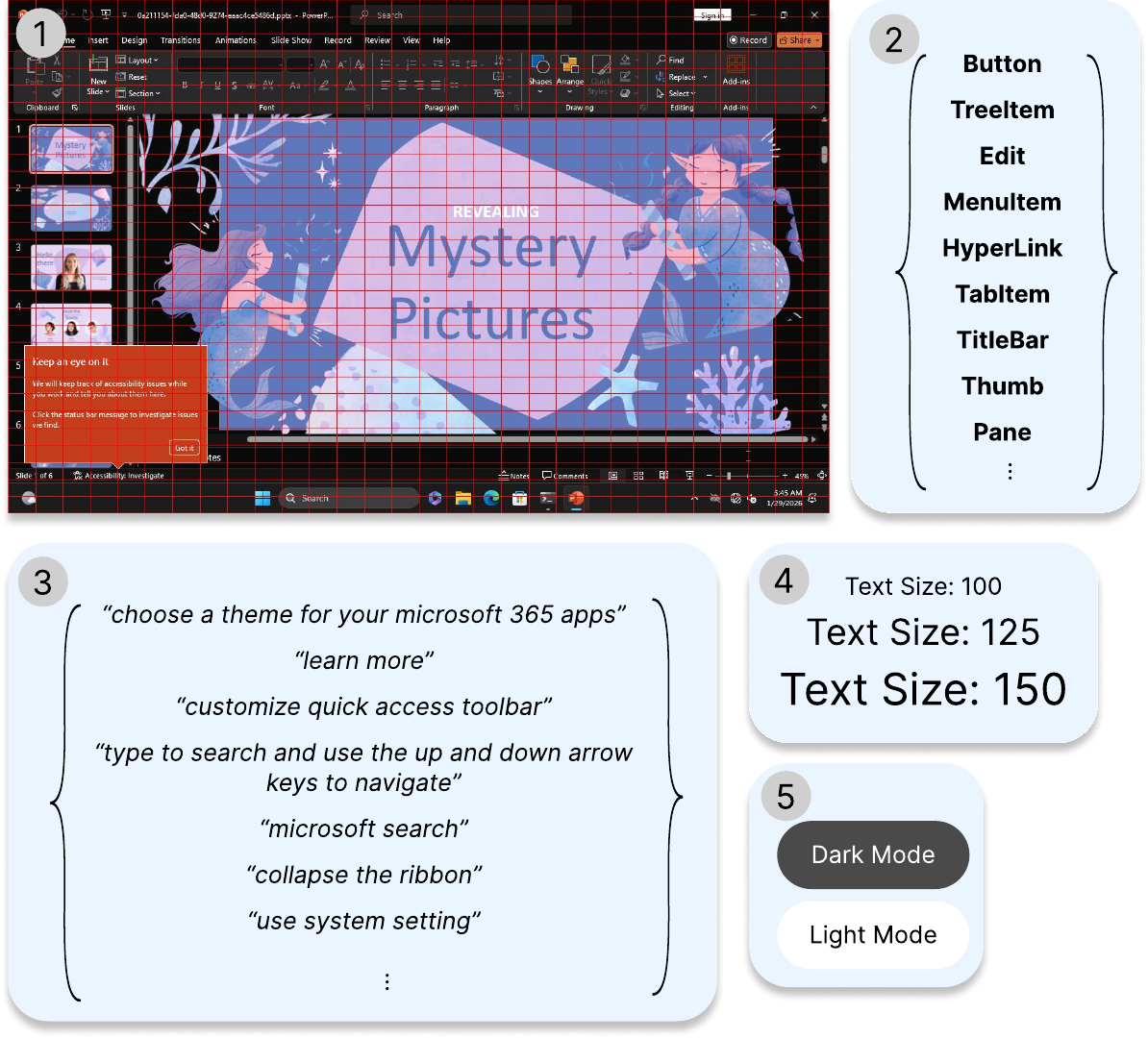}
    \caption{Discrete feature universes.}
\end{subfigure}\hfill
\begin{subfigure}[t]{0.63\linewidth}
    \centering
    \includegraphics[width=\linewidth]{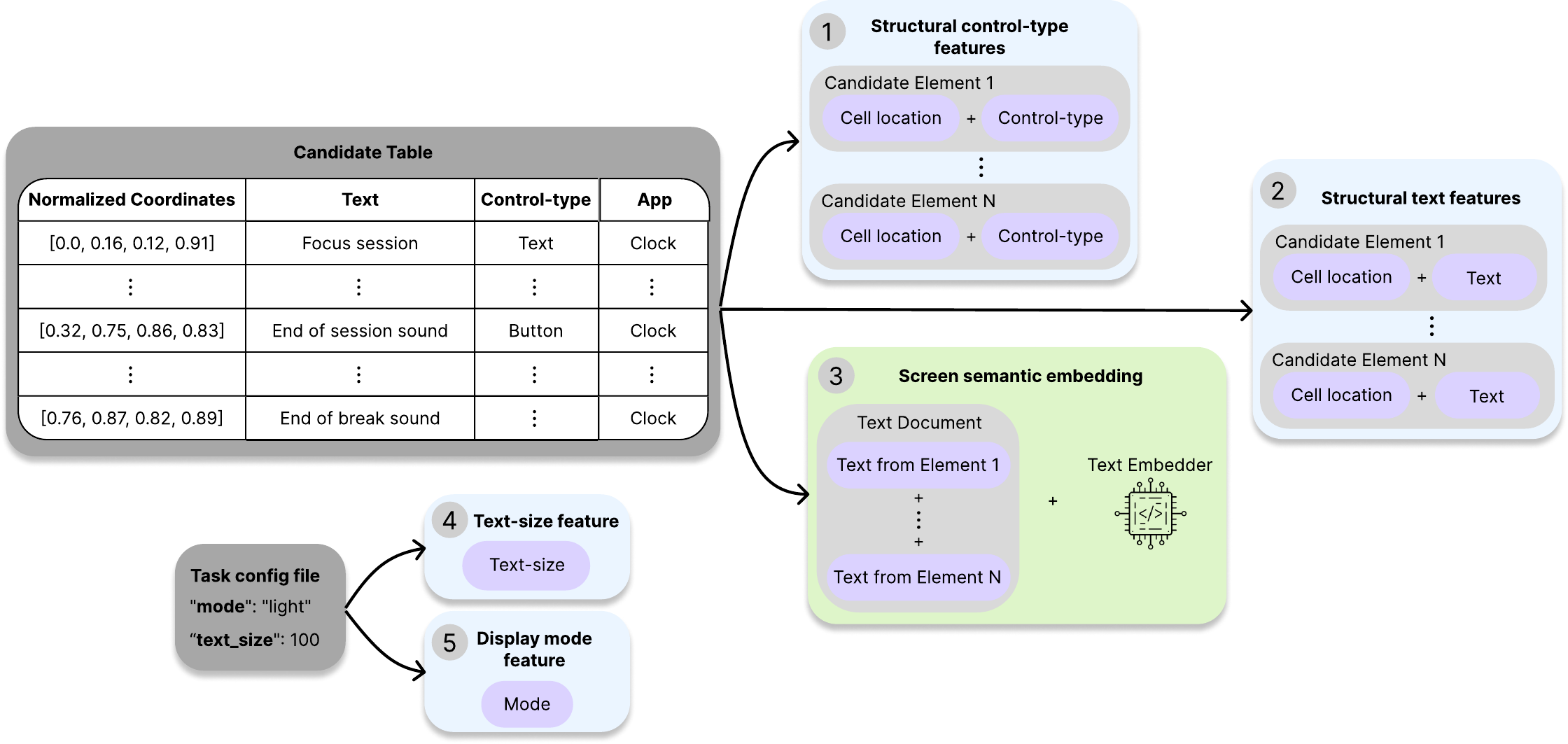}
    \caption{Feature extraction.}
\end{subfigure}
\caption{The active representation combines spatial structural atoms, normalized text, a dense semantic vector, and screen-level metadata.}
\label{fig:features}
\end{figure}

\subsection{Hybrid retrieval and bounded verification}

Given a query $S_q$ and an accumulated corpus $\mathcal{D}$, exact display-mode and text-size filters first define an eligible pool. Within that pool, BM25 ranks serialized sparse atoms and HNSW approximate nearest-neighbor search ranks dense vectors. Reciprocal Rank Fusion combines the two lists, and the fused top 200 proceed to verification. This two-stage design separates high-recall candidate generation from the stricter decision that alters graph identity.

For verification, let $T_{\mathrm{ct}}(S)$ and $T_{\mathrm{txt}}(S)$ be the sets of cell--control and cell--text atoms. We compute their Jaccard overlaps separately and use the bounded score
\begin{equation}
J_{\mathrm{sparse}}(S_q,S)=0.5J\!\left(T_{\mathrm{ct}}(S_q),T_{\mathrm{ct}}(S)\right)
+0.5J\!\left(T_{\mathrm{txt}}(S_q),T_{\mathrm{txt}}(S)\right),
\end{equation}
with merge threshold $\tau=0.93$. A query joins the highest-ranked candidate that passes this test; otherwise it becomes a new graph state. Dense similarity can therefore improve recall but cannot by itself merge two observations. This choice makes the definition of state identity explicit and keeps its downstream effects on novelty, visits, and ambiguity auditable. The complete pipeline is diagrammed in \cref{fig:screen_retrieval}; sensitivity to $\tau$ and exact-matching alternatives is reported in \cref{app:sensitivity}.

\section{One-step PUCT action selection}
\label{sec:puct}

At state $s$, the system enumerates executable action signatures and selects
\begin{equation}
\sigma^*=\arg\max_{\sigma}\left[Q(s,\sigma)+c_pP(\sigma\mid s)
\frac{\sqrt{\sum_{\tilde\sigma}N(s,\tilde\sigma)}}{1+N(s,\sigma)}\right],
\label{eq:puct}
\end{equation}
with $c_p=1.0$. The selected action is executed once in the VM. Relative to the graph snapshot immediately before execution, its observed reward is
\begin{equation}
r(s,\sigma,s')=\lambda_{\mathrm{state}}\ind[s'\notin V_t]
+\lambda_{\mathrm{edge}}\ind[(s,\sigma,s')\notin E_t]
+\lambda_{\mathrm{amb}}\pospart{u(s)-u(s')},
\label{eq:reward}
\end{equation}
where all three coefficients equal 1.0. The shared store increments $N(s,\sigma)$ and updates $Q(s,\sigma)$ as the running mean of this immediate reward. There are no simulated rollouts, no tree backpropagation, and no discounted return; the configured $\gamma=1.0$ is therefore unused.

This update has a useful distributed interpretation. Workers that reach the same merged state contribute to the same action counts and empirical values even when they arrived through different trajectories. Frequently tried low-reward signatures are gradually suppressed by their visit denominator, whereas underexplored actions retain a larger exploration bonus. Because the destination is observed in the real VM, the statistics incorporate actual application behavior rather than a learned transition model. Their reliability nevertheless remains coupled to state merging and action canonicalization, which motivates the exact definitions in \cref{sec:formulation,sec:retrieval}.

Unless otherwise stated, the deterministic proposal score is
\begin{equation}
\tilde P(\sigma\mid s)=1+0.2\left|\mathrm{tokens}(g)\cap\mathrm{tokens}(\mathrm{label}(\sigma))\right|
+0.15\ind[\mathrm{editable}(\sigma)\wedge\mathrm{type}(\sigma)],
\label{eq:heuristic_prior}
\end{equation}
normalized over current candidates, where $g$ is the exploration instruction. We also evaluate uniform and learned diffusion priors. The proposal and the accumulated PUCT statistics make separable contributions in \cref{sec:results}.

\section{Proposal priors from exploration traces}
\label{sec:trace_learning}

The PUCT graph-bandit runs produce a large corpus of short interaction traces that expose aliasing, loops, and failure modes. The main contribution is the exploration pipeline itself, but the traces also support lightweight proposal priors over executable action signatures conditioned on the current screen. A proposal changes only $P(\sigma\mid s)$ in \cref{eq:puct}; it is not treated as a standalone planner with a separate control objective, and the graph-bandit remains responsible for frontier-sensitive selection and ambiguity reduction.

The original corpus-building profile uses \texttt{uniform\_prior@1}. An optional deterministic \texttt{heuristic@1} profile ranks plausible signatures using the local structural and lexical cues in \cref{eq:heuristic_prior}; the matched three-application selector study uses this profile for both PUCT and greedy selection so that the selector is isolated. We additionally train a conditional diffusion proposal from retained traces. Comparing uniform, deterministic, and learned proposals lets us separate local proposal quality from the shared graph statistics accumulated by search.

Each worker records the screen representation, candidate table, canonical action, successor identity, and exploration statistics. Post-processing removes repeated cycles and near-no-op segments, then tokenizes retained actions by operation, target role and text, quantized position, and optional argument. The complete schema and marginal frequencies appear in \cref{appendix:action_tokenization,app:action_token_marginals}. A separate downstream experiment later uses the same traces for screen-conditioned action prediction; future screens are used only by training auxiliaries, and its multi-step metrics remain teacher-forced offline diagnostics.

\section{OS exploration setup}
\label{sec:setup}

\begin{figure*}[t]
\centering
\includegraphics[width=0.98\linewidth]{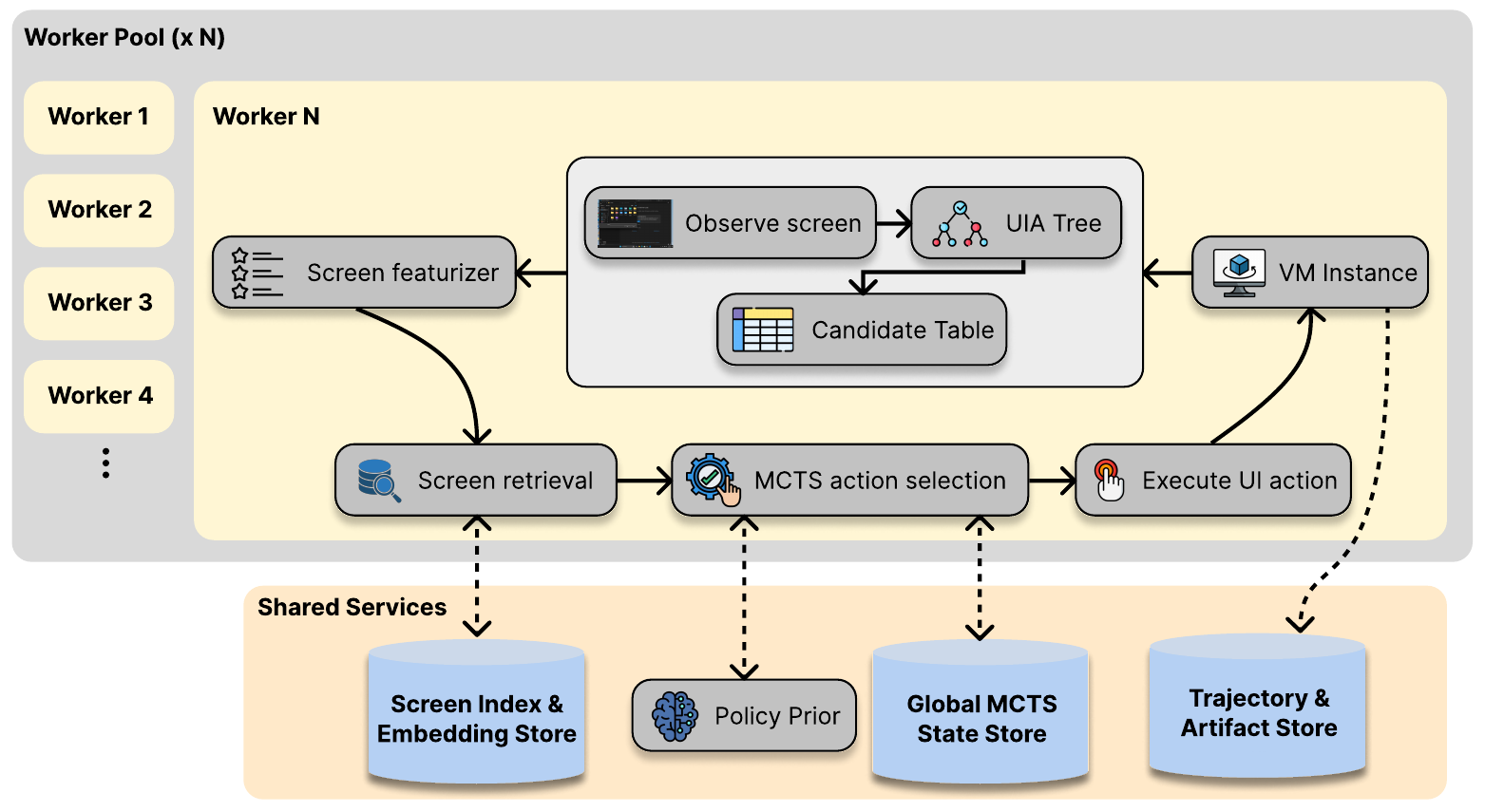}
\caption{Original distributed OS-exploration system overview. In the terminology used throughout this paper, the diagram's legacy ``MCTS action selection'' and ``Global MCTS State Store'' labels denote the one-step PUCT selector and its shared graph-statistics store, respectively; the reported system performs neither simulated rollouts nor tree backpropagation.}
\label{fig:system}
\end{figure*}

We evaluate \algacro{} as a large-scale desktop GUI exploration and data-generation system. For each application, we report VM-hours, cross-application captures, total screenshots, post-processed unique observations, merged graph states, and extracted three-step traces. The architecture in \cref{fig:system} separates replicated VM workers from the shared index, PUCT statistics, and artifact stores.

\paragraph{Large-scale corpus.}
Exploration runs begin from clean Windows VM snapshots, followed by a full reset between runs so that latent application state does not accumulate across trials. Seeded content, light or dark display mode, application size, and window placement vary across configurations to expose different observable regions. The large-scale collection totals 5,524 recorded VM-hours across 11 applications (\cref{tab:corpus}).

\paragraph{Action space and cross-application capture.}
Mouse clicks and keyboard typing are instantiated from the current UIA candidate table. Clicks target candidate elements; type actions are available only for text-editable targets and retain their exact argument in the action signature. Candidates are restricted to elements attributed to the target application. When an interaction opens a system dialog, file picker, authentication surface, or auxiliary application, the transition is logged as a cross-application capture rather than silently folded into the target-app state graph.

\paragraph{Trajectory extraction.}
Each run yields a raw interaction trace. Post-processing removes segments dominated by short loops and near-no-op actions, then extracts short windows that exhibit sustained observable progression. Structured UI metadata is retained alongside screenshots so that task descriptions, grounded element labels, and counterfactual training pairs can be regenerated offline without rerunning the VM exploration. Prompt templates and action-token construction appear in \cref{appendix:processing_prompts,appendix:action_tokenization}.

\paragraph{Deduplication hyperparameters.}
The similarity threshold determines whether a new observation joins an existing state or is indexed as new. The two component weights balance structural control-role evidence against spatial text evidence in the bounded duplicate score. We use the same values throughout the reported corpus and controlled selector runs.

\begin{table}[t]
\centering
\small
\caption{State-matching hyperparameters used for OS exploration.}
\label{tab:state_matching_hyperparameters}
\begin{tabular}{lc}
\toprule
Hyperparameter & Value\\
\midrule
Similarity threshold $\tau$ & 0.93\\
Cell$\times$control-role Jaccard weight & 0.5\\
Cell$\times$text Jaccard weight & 0.5\\
\bottomrule
\end{tabular}
\end{table}

\paragraph{Matched policy suite.}
The primary controlled study uses 50 verified replay starts in each of Notepad, File Explorer, and LibreOffice Calc, stratified as 17 low-, 16 medium-, and 17 high-ambiguity starts. A deterministic prefix replays each state from a clean snapshot. Conditions share starts, candidates, seeds, 50-action budgets, and all non-ablated settings. ``New states'' and ``new edges'' count coverage beyond the initialized replay artifact. We report means over the 50 starts per application. The original three-start Notepad diagnostic and its standard deviations are retained in \cref{app:notepad_diagnostic}, but are not used as cross-application evidence.

\paragraph{Metrics.}
Final counts and area under the cumulative discovery curve (AUC) measure additional state and edge coverage. Revisit rate is the fraction of steps returning to an episode-visited state. Loop/stall rate is the fraction belonging to a repeated short cycle or a no-progress run. High-ambiguity occupancy is the fraction of visited states above the fixed ambiguity threshold used for the replay suite. Net ambiguity change is the final minus initial $u(s)$.

\paragraph{Reactive baseline interface.}
Reactive models receive the instruction, active-window metadata, exploration counters, and at most 60 UIA candidates with ID, label, role, rectangle, clickability, and editability. They emit one validated JSON action from \{\texttt{click}, \texttt{type}, \texttt{wait}\}. Full prompting, sampling, and context details are in \cref{app:reproducibility}.

\section{Large-scale OS exploration results}
\label{sec:corpus_results}

\subsection{Overall performance}

\begin{table*}[t]
\centering
\scriptsize
\setlength{\tabcolsep}{3.5pt}
\caption{Large-scale OS exploration corpus. ``Unique obs.'' are post-processed screenshot identities; ``graph states'' apply the final bounded merge. The last column counts extracted length-three traces.}
\label{tab:corpus}
\resizebox{\textwidth}{!}{%
\begin{tabular}{lrrrrrrr}
\toprule
Application & VM hours & Cross-app & Screenshots & Unique obs. & Graph states & State rate (\%) & 3-step traces\\
\midrule
Clock & 214 & 1 & 42,500 & 1,366 & 1,366 & 3.21 & 3,701\\
File Explorer & 366 & 8 & 52,763 & 3,985 & 3,985 & 7.55 & 2,662\\
Microsoft Excel & 663 & 6 & 86,025 & 2,585 & 2,585 & 3.00 & 4,324\\
Microsoft Word & 1,377 & 1 & 213,613 & 8,759 & 8,759 & 4.10 & 10,777\\
Microsoft PowerPoint & 1,809 & 1 & 466,115 & 8,124 & 8,124 & 1.74 & 15,834\\
Notepad & 110 & 1 & 30,445 & 314 & 314 & 1.03 & 612\\
Settings & 460 & 20 & 53,037 & 2,310 & 2,310 & 4.36 & 13,457\\
VS Code & 250 & 1 & 39,533 & 1,944 & 1,944 & 4.92 & 1,995\\
LibreOffice Writer & 92 & 2 & 7,840 & 2,420 & 587 & 7.49 & 924\\
LibreOffice Calc & 85 & 1 & 7,841 & 2,642 & 614 & 7.83 & 396\\
LibreOffice Math & 98 & 1 & 7,694 & 5,453 & 558 & 7.25 & 1,088\\
\midrule
\textbf{Overall} & \textbf{5,524} & \textbf{43} & \textbf{1,007,406} & \textbf{39,902} & \textbf{31,146} & \textbf{3.09} & \textbf{55,770}\\
\bottomrule
\end{tabular}
}
\end{table*}

The 39,902 unique observations in \cref{tab:corpus} are post-processed screenshot identities before the final bounded state merge; applying the active representation and threshold produces 31,146 graph states. State discovery rates range from 1.03\% in Notepad to 7.83\% in LibreOffice Calc, reflecting differences in interface variability and reachable seeded workflows. The 43 cross-application captures show that target-restricted actions can still surface operating-system dialogs and auxiliary applications, while the 55,770 extracted traces quantify the short interaction windows retained for subsequent analysis.

\begin{figure*}[t]
\centering
\begin{subfigure}[t]{0.48\linewidth}
    \centering
    \includegraphics[width=\linewidth,height=0.20\textheight,keepaspectratio]{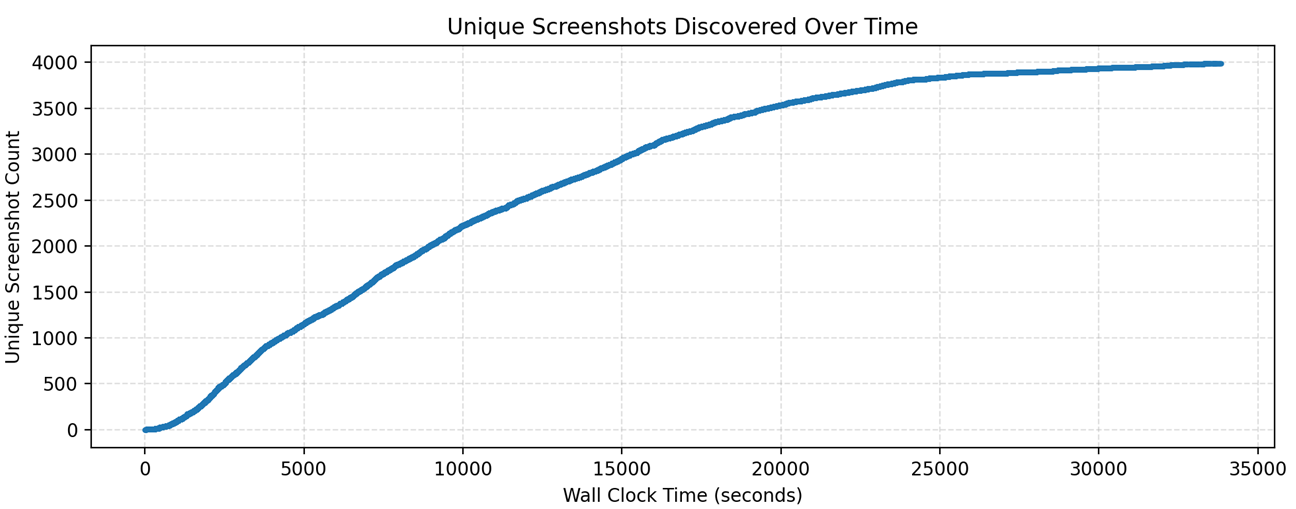}
    \caption{Cumulative unique observations over wall-clock time.}
    \label{fig:time_vs_number_unique_states}
\end{subfigure}\hfill
\begin{subfigure}[t]{0.48\linewidth}
    \centering
    \includegraphics[width=\linewidth,height=0.20\textheight,keepaspectratio]{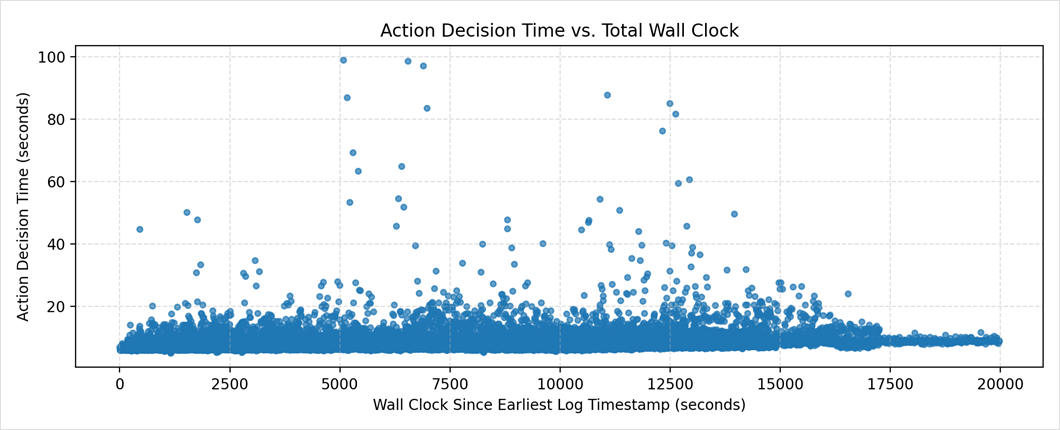}
    \caption{Per-step decision latency as the corpus grows.}
    \label{fig:action_decision_time_vs_wall_clock}
\end{subfigure}

\vspace{0.5em}
\begin{subfigure}[t]{0.48\linewidth}
    \centering
    \includegraphics[width=\linewidth,height=0.20\textheight,keepaspectratio]{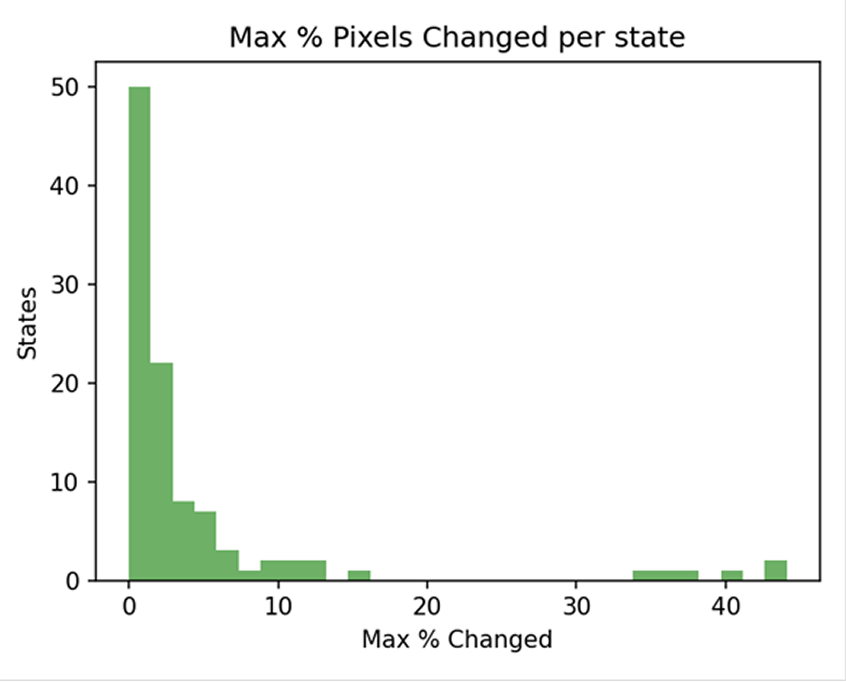}
    \caption{Maximum changed-pixel fraction within merged states.}
    \label{fig:pixels_changed_per_state}
\end{subfigure}\hfill
\begin{subfigure}[t]{0.48\linewidth}
    \centering
    \includegraphics[width=\linewidth,height=0.20\textheight,keepaspectratio]{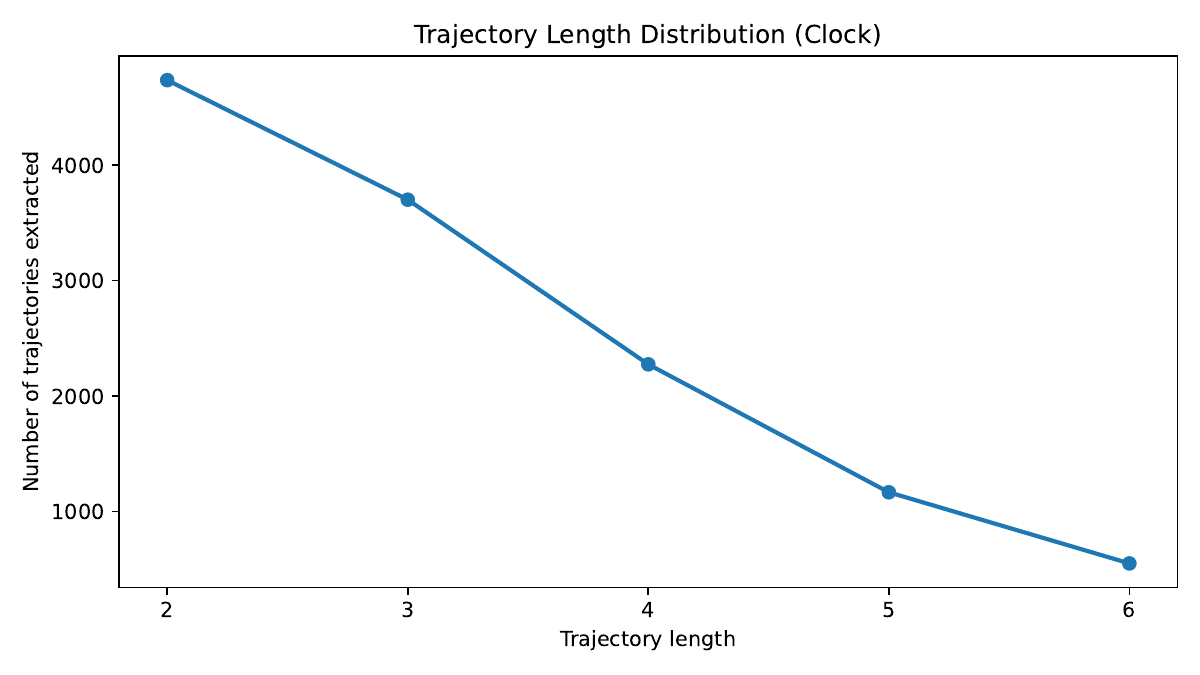}
    \caption{Extracted Clock trajectories by length.}
    \label{fig:trajectory_length_line_graph}
\end{subfigure}
\caption{Large-scale OS-exploration diagnostics. The plots restore the temporal discovery, scaling, state-coherence, and trace-extraction views used to interpret the corpus.}
\label{fig:corpus_diagnostics}
\end{figure*}

\paragraph{State discovery over time.}
The cumulative curve in \cref{fig:time_vs_number_unique_states} rises rapidly while workers traverse high-branching UI regions, then gradually flattens as exploration encounters more revisits and near-duplicates. The curve is therefore useful alongside the final corpus totals: it distinguishes sustained discovery from a system that produces most of its coverage in a brief initial phase.

\paragraph{Efficiency, state coherence, and trace extraction.}
Per-step action-decision latency remains stable over the collection run in \cref{fig:action_decision_time_vs_wall_clock}, with isolated spikes consistent with transient VM or application delays rather than steady growth with index size. The changed-pixel distribution in \cref{fig:pixels_changed_per_state} is concentrated near zero, providing an independent visual-coherence diagnostic for merged states. Finally, \cref{fig:trajectory_length_line_graph} shows the expected short-trace skew after filtering loop-heavy and near-no-op segments.

\paragraph{Prior-guided corpus exploration.}
In the original corpus-building comparison, replacing uniform action weights with the deterministic heuristic prior raises unique-state discovery from 1.74\% to 4.10\%. This comparison is descriptive because it comes from the large-scale generation profile rather than the matched replay suite; the controlled learned-prior and fixed-prior selector comparisons are reported separately in \cref{app:notepad_diagnostic,tab:cross_app_selector}.

\begin{table}[t]
\centering
\small
\caption{Proposal prior effect in the original corpus-building comparison.}
\label{tab:corpus_prior}
\begin{tabular}{lc}
\toprule
Proposal prior & Unique-state discovery rate (\%)\\
\midrule
Deterministic heuristic & 4.10\\
Uniform & 1.74\\
\bottomrule
\end{tabular}
\end{table}

\section{Exploration under uncertainty: policy evaluation}
\label{sec:results}

Large-scale corpus statistics establish coverage but do not isolate the action selector. We therefore evaluate \algacro{} as a policy problem under uncertainty: useful behavior should expand the discovered frontier while avoiding repeated commitment to already known or low-yield branches. The primary replay-start suite measures additional state and edge coverage, ambiguity trajectories, revisits, and stalls under matched executable candidates and budgets.

\subsection{Matched cross-application selector comparison}

\Cref{tab:cross_app_selector} holds proposal probabilities fixed and changes only greedy selection versus the PUCT statistics in \cref{eq:puct}. PUCT improves new-state and new-edge AUC in every application. At the macro level, new states rise from 0.100 to 1.573 and state AUC from 4.820 to 50.487, while revisit falls from 0.9980 to 0.9678. The high absolute revisit rates reflect difficult starts initialized from an existing replay artifact; only additional coverage receives credit.

\begin{table*}[t]
\centering
\small
\setlength{\tabcolsep}{4.2pt}
\caption{Matched cross-application selector comparison. Both rows per application use the same deterministic heuristic prior. Values are means over 50 starts.}
\label{tab:cross_app_selector}
\begin{tabular}{llrrrrr}
\toprule
Application & Selector & New states @ 50 & State AUC & Edge AUC & Net $\Delta u$ & Revisit\\
\midrule
Notepad & PUCT & 1.000 & 32.900 & 361.960 & 0.0608 & 0.9796\\
 & Greedy & 0.240 & 11.560 & 78.500 & -0.0505 & 0.9951\\
File Explorer & PUCT & 3.140 & 98.080 & 1,076.400 & 0.0351 & 0.9357\\
 & Greedy & 0.040 & 1.920 & 55.640 & 0.0691 & 0.9992\\
LibreOffice Calc & PUCT & 0.580 & 20.480 & 709.720 & -0.0711 & 0.9882\\
 & Greedy & 0.020 & 0.980 & 49.940 & 0.0182 & 0.9996\\
\midrule
Macro-average & PUCT & \textbf{1.573} & \textbf{50.487} & \textbf{716.027} & 0.0083 & \textbf{0.9678}\\
 & Greedy & 0.100 & 4.820 & 61.360 & 0.0123 & 0.9980\\
\bottomrule
\end{tabular}
\end{table*}

Proposal quality remains important. On the original matched Notepad diagnostic, replacing a uniform prior with a learned diffusion prior raises PUCT frontier states from 10.00 to 14.33 and state AUC from 253.33 to 406.67, while reducing revisit from 0.796 to 0.708 and loop/stall from 0.440 to 0.253 (\cref{tab:notepad_prior}). Thus learned proposals and shared PUCT statistics are complementary rather than interchangeable.

\subsection{Objective and component ablations}

\Cref{tab:component_ablation} uses the same three-application starts and heuristic prior except where named. Relative to novelty-only, the full objective raises state AUC by 8.9\% and modestly lowers revisit and loop/stall, but novelty-only has higher edge AUC and lower high-ambiguity occupancy. Prior-only and uniform-prior variants find more raw states but also spend more time in high-ambiguity states. These results support a coverage--ambiguity trade-off, not uniform dominance of the full reward. Exact matching changes the definition of state identity, so its larger raw count is not interpreted as better semantic coverage.

\begin{table*}[t]
\centering
\small
\setlength{\tabcolsep}{4pt}
\caption{Three-application macro-average component ablations under matched starts, candidates, seeds, and budgets.}
\label{tab:component_ablation}
\begin{tabular}{lrrrrrr}
\toprule
Configuration & New states & State AUC & Edge AUC & High-$u$ occ. & Revisit & Loop/stall\\
\midrule
Full: novelty + ambiguity & 1.573 & 50.487 & 716.027 & 0.6568 & 0.9678 & 0.8593\\
Novelty only & 1.427 & 46.347 & 901.533 & 0.4907 & 0.9705 & 0.8828\\
Ambiguity only & 1.047 & 34.300 & 761.987 & 0.6818 & 0.9786 & 0.9000\\
Prior only & 2.233 & 71.867 & 982.900 & 0.7068 & 0.9544 & 0.8153\\
Full + uniform prior & 4.733 & 132.527 & 1,094.720 & 0.7983 & 0.9028 & 0.6937\\
Full + exact matching & 2.100 & 67.847 & 997.240 & 0.6038 & 0.9570 & 0.8651\\
\bottomrule
\end{tabular}
\end{table*}

One-factor sensitivity results are reported in \cref{app:sensitivity}. Increasing $\tau$ reduces false merges but fragments states, $\kappa$ has a comparatively modest effect around 20, and $u_0$ is a meaningful operating choice. Pair F1 uses persisted UIA IDs as a matcher diagnostic, not independent semantic ground truth.

\subsection{Controlled ambiguity and probe validation}
\label{sec:controlled}

The same visible Notepad state was reset with clipboard value \texttt{alpha} or \texttt{beta}; Paste is deterministic within either variant but differs when variants are pooled. An engineered delayed clipboard update instead varies within one fixed condition. \Cref{tab:causal_control} shows the intended decomposition: known hidden state produces between-condition dispersion, while timing noise remains within condition. The scalar $u(s)$ alone does not perform this causal attribution; repeated condition labels are required.

\begin{table}[t]
\centering
\small
\setlength{\tabcolsep}{4pt}
\caption{Controlled matched-action dispersion.}
\label{tab:causal_control}
\begin{tabular}{lrrrr}
\toprule
Source & Trials & Pooled & Within & Between\\
\midrule
Hidden clipboard variants & 20 & 1.000 & 0.000 & 1.000\\
Asynchronous race & 10 & 0.971 & 0.971 & 0.000\\
\bottomrule
\end{tabular}
\end{table}

We then test probe utility with 20 runs per condition. A reversible Paste discloses which hidden clipboard token is present; Undo restores the document before the policy types the corresponding commitment token. Opening and closing the File menu reaches an equally repeatable, low-dispersion destination but reveals nothing. Paste raises commitment accuracy from 0.500 to 1.000; menu navigation leaves it at 0.500 (\cref{tab:probe_control}). This is why lower destination dispersion alone is insufficient to call an action a probe.

\begin{table}[t]
\centering
\small
\caption{Probe-versus-escape control, 20 runs per condition.}
\label{tab:probe_control}
\begin{tabular}{lrrr}
\toprule
Condition & Disclosure & Destination dispersion & Commitment accuracy\\
\midrule
No probe & 0.000 & -- & 0.500\\
Reversible Paste + Undo & 1.000 & 0.000 & 1.000\\
File-menu navigation & 0.000 & 0.000 & 0.500\\
\bottomrule
\end{tabular}
\end{table}

An audit of 1,588 repeated state--action groups across the three applications provides a less synthetic check: nearly all measured dispersion lies between recorded contexts rather than within a fixed context (\cref{tab:context_decomposition}). This supports the proxy's usefulness while preserving the caveat that unrecorded timing and merge noise can still contribute.

\subsection{Observation representation and downstream use}

The OCR audit re-encodes the same 1,200 screenshots with UIA features or RapidOCR 1.4.4 using PP-OCRv4 \citep{rapidocr2025,paddleocr2023}. UIA obtains same-state pair F1 0.740 versus 0.646 for OCR, while OCR obtains recall@10 0.409 versus 0.381, false-merge rate 0.007 versus 0.010, and successor consistency 0.958 versus 0.892. Because reference labels are existing \algacro{} IDs, the comparison favors UIA; nevertheless, OCR remains a viable input to the same retrieval and grouping pipeline (\cref{app:representation_audit}).

For downstream evaluation, we slice Windows Settings traces into 22,028 three-step screen--action windows with 8,118 task descriptions; 93.99\% contain a state change. We compare the same step-conditioned diffusion architecture trained with and without counterfactual pairs that hold task text fixed while changing the screen and target action. The split is task-template-disjoint and uses 1,000 frozen prompts. Future screens are used only for training auxiliaries, not action prediction.

\begin{table*}[t]
\centering
\small
\setlength{\tabcolsep}{4pt}
\caption{Downstream screen-conditioned policy results. Sensitivities are accuracy drops after screen shuffling. Three-step success and recovery are teacher-forced offline proxies, not live GUI success.}
\label{tab:downstream}
\begin{tabular}{lrrrrr}
\toprule
Training setup & Step match & Screen sens. & Change sens. & Offline 3-step & Recovery\\
\midrule
With counterfactual pairs & 52.60\% & 22.50 pts & 23.45 pts & 1.74\% & 24.35\%\\
Without counterfactual pairs & 53.10\% & 18.80 pts & 18.95 pts & 0.85\% & 21.68\%\\
\bottomrule
\end{tabular}
\end{table*}

Counterfactual supervision preserves one-step match while increasing screen sensitivity by 3.70 points, state-change sensitivity by 4.50 points, offline three-step success from 0.85\% to 1.74\%, and recovery from 21.68\% to 24.35\%. The policy changes its prediction on 66.52\% of same-task/different-screen pairs. These results show that the traces can train screen-dependent behavior; they do not establish live end-to-end task completion.

\section{Limitations and broader impacts}
\label{sec:limitations}

The ambiguity score is an empirical structural proxy, not a semantic posterior. It detects unreliable matched-action outcomes but cannot determine from one observation whether dispersion comes from hidden workflow state, asynchronous execution, stochasticity, or retrieval and merge errors. The controlled decomposition requires repeated trials or condition labels that will often be unavailable in deployment. Likewise, a transition to lower $u(s)$ is not necessarily informative; the probe experiment evaluates disclosure through a later commitment, whereas the general selector optimizes only the proxy in \cref{eq:reward}.

The primary matched policy study covers three applications and a 50-action budget, despite the corpus spanning 11 applications. UIA metadata is effective for conventional desktop controls but can be missing for custom-rendered, canvas-based, or continuously changing interfaces. The OCR audit tests an alternative representation offline on stored screenshots; it does not establish equivalent online performance. The PUCT graph-bandit is one-step and therefore cannot by itself reason about delayed information value or irreversible multi-action consequences. Its quality remains coupled to candidate generation, replay fidelity, the retrieval index, and action-signature design.

The downstream metrics are teacher-forced offline proxies and the absolute three-step success rates are low; live task completion, safety, and recovery remain open. Large-scale GUI exploration can also expose sensitive text or execute destructive actions if deployed outside a sandbox. Our experiments use isolated clean-snapshot VMs, restrict candidates to the target application, and log cross-application transitions. Any external deployment or artifact release should additionally redact secrets, require explicit authorization, and gate irreversible actions. These safeguards limit direct user harm but do not eliminate dual-use risks from more capable autonomous computer control.

\section{Conclusion}

\algacro{} treats desktop exploration as the joint problem of building a reusable state graph and recognizing when matched actions have unreliable outcomes. Its representation and hybrid retrieval system provide stable state identity at corpus scale, while the shared PUCT graph-bandit turns observed frontier and ambiguity signals into coordinated action statistics across workers. Matched experiments show that those statistics improve additional coverage over greedy use of the same prior, and the objective and proposal ablations reveal a genuine coverage--ambiguity trade-off rather than a universally dominant configuration. Controlled tests clarify both the value and limits of the dispersion proxy. Finally, the learned-prior, OCR, and downstream-policy studies show how the resulting graph and traces can support other observation encoders and screen-grounded models. Explicit state identity, proposal quality, and uncertainty-aware statistics are therefore complementary tools for deciding when a desktop agent should continue exploring before it commits.

\bibliography{reference}

@article{zhou2023webarena, title={WebArena: A Realistic Web Environment for Building Autonomous Agents}, author={Zhou, Shuyan and Xu, Frank F. and Zhu, Hao and Zhou, Xuhui and Lo, Robert and Sridhar, Abishek and Cheng, Xianyi and Ou, Tianyue and Bisk, Yonatan and Fried, Daniel and Alon, Uri and Neubig, Graham}, journal={arXiv:2307.13854}, year={2023}}

@article{deng2023mind2web, title={Mind2Web: Towards a Generalist Agent for the Web}, author={Deng, Xiang and Gu, Yu and Zheng, Boyuan and Chen, Shijie and Stevens, Samuel and Wang, Boshi and Sun, Huan and Su, Yu}, journal={arXiv:2306.06070}, year={2023}}

@article{liu2018miniwob, title={Reinforcement Learning on Web Interfaces using Workflow-Guided Exploration}, author={Liu, Evan Z. and others}, journal={arXiv:1802.08802}, year={2018}}

@article{kagaya2024rap, title={RAP: Retrieval-Augmented Planning with Contextual Memory for Multimodal Agents}, author={Kagaya, T. and others}, journal={arXiv:2402.03610}, year={2024}}

@article{chen2021decisiontransformer,
  title   = {Decision Transformer: Reinforcement Learning via Sequence Modeling},
  author  = {Chen, Lili and Lu, Kevin and Rajeswaran, Aravind and Lee, Kimin and Grover, Aditya and Laskin, Michael and Abbeel, Pieter and Srinivas, Aravind and Mordatch, Igor},
  journal = {arXiv preprint arXiv:2106.01345},
  year    = {2021}
}

@article{silver2017mastering,
  title={Mastering the game of Go without human knowledge},
  author={Silver, David and Schrittwieser, Julian and Simonyan, Karen and Antonoglou, Ioannis and Huang, Aja and Guez, Arthur and Hubert, Thomas and Baker, Lucas and Lai, Matthew and Bolton, Adrian and others},
  journal={Nature},
  volume={550},
  number={7676},
  pages={354--359},
  year={2017},
  publisher={Nature Publishing Group}
}

@article{silver2018general,
  title={A general reinforcement learning algorithm that masters chess, shogi, and Go through self-play},
  author={Silver, David and Hubert, Thomas and Schrittwieser, Julian and Antonoglou, Ioannis and Lai, Matthew and Guez, Arthur and Lanctot, Marc and Sifre, Laurent and Kumaran, Dharshan and Graepel, Thore and others},
  journal={Science},
  volume={362},
  number={6419},
  pages={1140--1144},
  year={2018},
  publisher={American Association for the Advancement of Science}
}

@article{yao2022react,
  title   = {ReAct: Synergizing Reasoning and Acting in Language Models},
  author  = {Yao, Shunyu and Zhao, Jeffrey and Yu, Dian and Du, Nan and Shafran, Izhak and Narasimhan, Karthik and Cao, Yuan},
  journal = {arXiv preprint arXiv:2210.03629},
  year    = {2022},
  doi     = {10.48550/arXiv.2210.03629},
  url     = {https://arxiv.org/abs/2210.03629}
}

@article{ahn2022saycan,
  title   = {Do As I Can, Not As I Say: Grounding Language in Robotic Affordances},
  author  = {Ahn, Michael and Brohan, Anthony and Brown, Noah and Chebotar, Yevgen and Cortes, Omar and David, Byron and Finn, Chelsea and Fu, Chuyuan and Gopalakrishnan, Keerthana and Hausman, Karol and Herzog, Alex and Ho, Daniel and Hsu, Jasmine and Ibarz, Julian and Ichter, Brian and Irpan, Alex and Jang, Eric and Jeffrey, Kyle and Jesmonth, Sam and Joshi, Nikhil and Julian, Ryan and Kalashnikov, Dmitry and Levine, Sergey and Lu, Liyiming and Parada, Carolina and Rao, Kanishka and Rückert, Ulrich and Schneider, Jonas and Sheng, Kelvin and Sermanet, Pierre and Sriram, Srinivas and Stone, Austin and Tan, Tianhe and Tran, Huong and Vanhoucke, Vincent and Xia, Fei and Xu, Ted and Zeng, Alan},
  journal = {arXiv preprint arXiv:2204.01691},
  year    = {2022},
  url     = {https://arxiv.org/abs/2204.01691}
}

@article{shinn2023reflexion,
  title   = {Reflexion: Language Agents with Verbal Reinforcement Learning},
  author  = {Shinn, Noah and Cassano, Federico and Berman, Edward and Gopinath, Ashwin and Narasimhan, Karthik and Yao, Shunyu},
  journal = {arXiv preprint arXiv:2303.11366},
  year    = {2023},
  doi     = {10.48550/arXiv.2303.11366},
  url     = {https://arxiv.org/abs/2303.11366}
}

@article{zhou2023lats,
  title   = {Language Agent Tree Search Unifies Reasoning Acting and Planning in Language Models},
  author  = {Zhou, Andy and Yan, Kai and Shlapentokh-Rothman, Michal and Wang, Haohan and Wang, Yu-Xiong},
  journal = {arXiv preprint arXiv:2310.04406},
  year    = {2023},
  doi     = {10.48550/arXiv.2310.04406},
  url     = {https://arxiv.org/abs/2310.04406}
}

@inproceedings{janner2022diffuser,
  title     = {Planning with Diffusion for Flexible Behavior Synthesis},
  author    = {Janner, Michael and Du, Yilun and Tenenbaum, Joshua and Levine, Sergey},
  booktitle = {Proceedings of the 39th International Conference on Machine Learning},
  series    = {Proceedings of Machine Learning Research},
  volume    = {162},
  pages     = {9902--9915},
  year      = {2022},
  publisher = {PMLR},
  url       = {https://proceedings.mlr.press/v162/janner22a.html}
}

@inproceedings{chi2023diffusionpolicy,
  title     = {Diffusion Policy: Visuomotor Policy Learning via Action Diffusion},
  author    = {Chi, Cheng and Feng, Siyuan and Du, Yilun and Xu, Zhenjia and Cousineau, Eric and Burchfiel, Benjamin and Song, Shuran},
  booktitle = {Proceedings of Robotics: Science and Systems (RSS)},
  year      = {2023}
}

@article{xie2024osworld,
  title={OSWorld: Benchmarking Multimodal Agents for Open-Ended Tasks in Real Computer Environments},
  author={Xie, Tianbao and Zhang, Danyang and Chen, Jixuan and Li, Xiaochuan and Zhao, Siheng and Cao, Ruisheng and Hua, Toh Jing and Cheng, Zhoujun and Shin, Dongchan and Lei, Fangyu and Liu, Yitao and Xu, Yiheng and Zhou, Shuyan and Savarese, Silvio and Xiong, Caiming and Zhong, Victor and Yu, Tao},
  journal={arXiv preprint arXiv:2404.07972},
  year={2024},
  url={https://arxiv.org/abs/2404.07972}
}

@InProceedings{pathak2017curiosity,
  title     = {Curiosity-driven Exploration by Self-supervised Prediction},
  author    = {Pathak, Deepak and Agrawal, Pulkit and Efros, Alexei A. and Darrell, Trevor},
  booktitle = {Proceedings of the 34th International Conference on Machine Learning},
  pages     = {2778--2787},
  year      = {2017},
  editor    = {Precup, Doina and Teh, Yee Whye},
  volume    = {70},
  series    = {Proceedings of Machine Learning Research},
  month     = {06--11 Aug},
  publisher = {PMLR},
  url       = {https://proceedings.mlr.press/v70/pathak17a.html}
}

@Article{burda2018rnd,
  title         = {Exploration by Random Network Distillation},
  author        = {Burda, Yuri and Edwards, Harrison and Storkey, Amos J. and Klimov, Oleg},
  journal       = {arXiv preprint arXiv:1810.12894},
  year          = {2018},
  doi           = {10.48550/arXiv.1810.12894},
  url           = {https://arxiv.org/abs/1810.12894},
  eprint        = {1810.12894},
  archivePrefix = {arXiv},
  primaryClass  = {cs.LG}
}

@InProceedings{chrisman1992aliasing,
  title     = {Reinforcement Learning with Perceptual Aliasing: The Perceptual Distinctions Approach},
  author    = {Chrisman, Lonnie},
  booktitle = {Proceedings of the Tenth National Conference on Artificial Intelligence (AAAI-92)},
  pages     = {183--188},
  year      = {1992},
  editor    = {Swartout, William R.},
  publisher = {AAAI Press},
  url       = {https://dblp.org/rec/conf/aaai/Chrisman92}
}

@article{veiga2023activesensing,
  title   = {From Reactive to Active Sensing: A Survey on Information Gathering in Decision-Theoretic Planning},
  author  = {Veiga, Tiago and Renoux, Jennifer},
  journal = {ACM Computing Surveys},
  volume  = {55},
  number  = {13s},
  pages   = {280:1--280:22},
  year    = {2023},
  doi     = {10.1145/3583068},
  url     = {https://dl.acm.org/doi/10.1145/3583068}
}

@article{bajcsy2018activeperception,
  title   = {Revisiting Active Perception},
  author  = {Bajcsy, Ruzena and Aloimonos, Yiannis and Tsotsos, John K.},
  journal = {Autonomous Robots},
  volume  = {42},
  number  = {2},
  pages   = {177--196},
  year    = {2018},
  doi     = {10.1007/s10514-017-9615-3},
  url     = {https://doi.org/10.1007/s10514-017-9615-3}
}

@inproceedings{silver2010pomcp,
  title     = {Monte-Carlo Planning in Large POMDPs},
  author    = {Silver, David and Veness, Joel},
  booktitle = {Advances in Neural Information Processing Systems 23},
  pages     = {2164--2172},
  year      = {2010},
  publisher = {Curran Associates, Inc.},
  url       = {https://papers.nips.cc/paper/4031-monte-carlo-planning-in-large-pomdps}
}

@inproceedings{sekar2020plan2explore,
  title     = {Planning to Explore via Self-Supervised World Models},
  author    = {Sekar, Ramanan and Rybkin, Oleh and Daniilidis, Kostas and Abbeel, Pieter and Hafner, Danijar and Pathak, Deepak},
  booktitle = {Proceedings of the 37th International Conference on Machine Learning},
  series    = {Proceedings of Machine Learning Research},
  volume    = {119},
  pages     = {8583--8592},
  year      = {2020},
  publisher = {PMLR},
  url       = {https://proceedings.mlr.press/v119/sekar20a.html}
}

@misc{rapidocr2025,
  author       = {{RapidAI Team}},
  title        = {{RapidOCR}: OCR Toolbox},
  year         = {2025},
  note         = {Version 1.4.4},
  howpublished = {\url{https://pypi.org/project/rapidocr-onnxruntime/1.4.4/}}
}

@misc{paddleocr2023,
  author       = {{PaddleOCR Team}},
  title        = {{PP-OCRv4} Technical Report},
  year         = {2023},
  howpublished = {\url{https://paddlepaddle.github.io/PaddleOCR/v2.9/ppocr/blog/PP-OCRv4_introduction.html}}
}

\clearpage
\appendix

\section{Reproducibility details}
\label{app:reproducibility}

\subsection{Reported-run configuration}

\begin{table}[H]
\centering
\small
\caption{Retrieval and one-step PUCT configuration used for the reported runs.}
\label{tab:reported_configuration}
\begin{tabular}{ll}
\toprule
Component & Configuration\\
\midrule
Metadata filter & Display mode and text-size bin\\
Sparse retrieval & BM25 over spatial control/text atoms\\
Dense retrieval & \texttt{text-embedding-3-small} with HNSW ANN\\
Fusion & RRF, fused top 200\\
Verification & 0.5 control + 0.5 text Jaccard; $\tau=0.93$\\
PUCT & $c_p=1.0$\\
Reward & $\lambda_{\mathrm{state}}=\lambda_{\mathrm{edge}}=\lambda_{\mathrm{amb}}=1.0$\\
Ambiguity & $\kappa=20$, $u_0=0.5$\\
\bottomrule
\end{tabular}
\end{table}

\begin{figure}[H]
    \centering
    \includegraphics[width=0.82\linewidth,height=0.24\textheight,keepaspectratio]{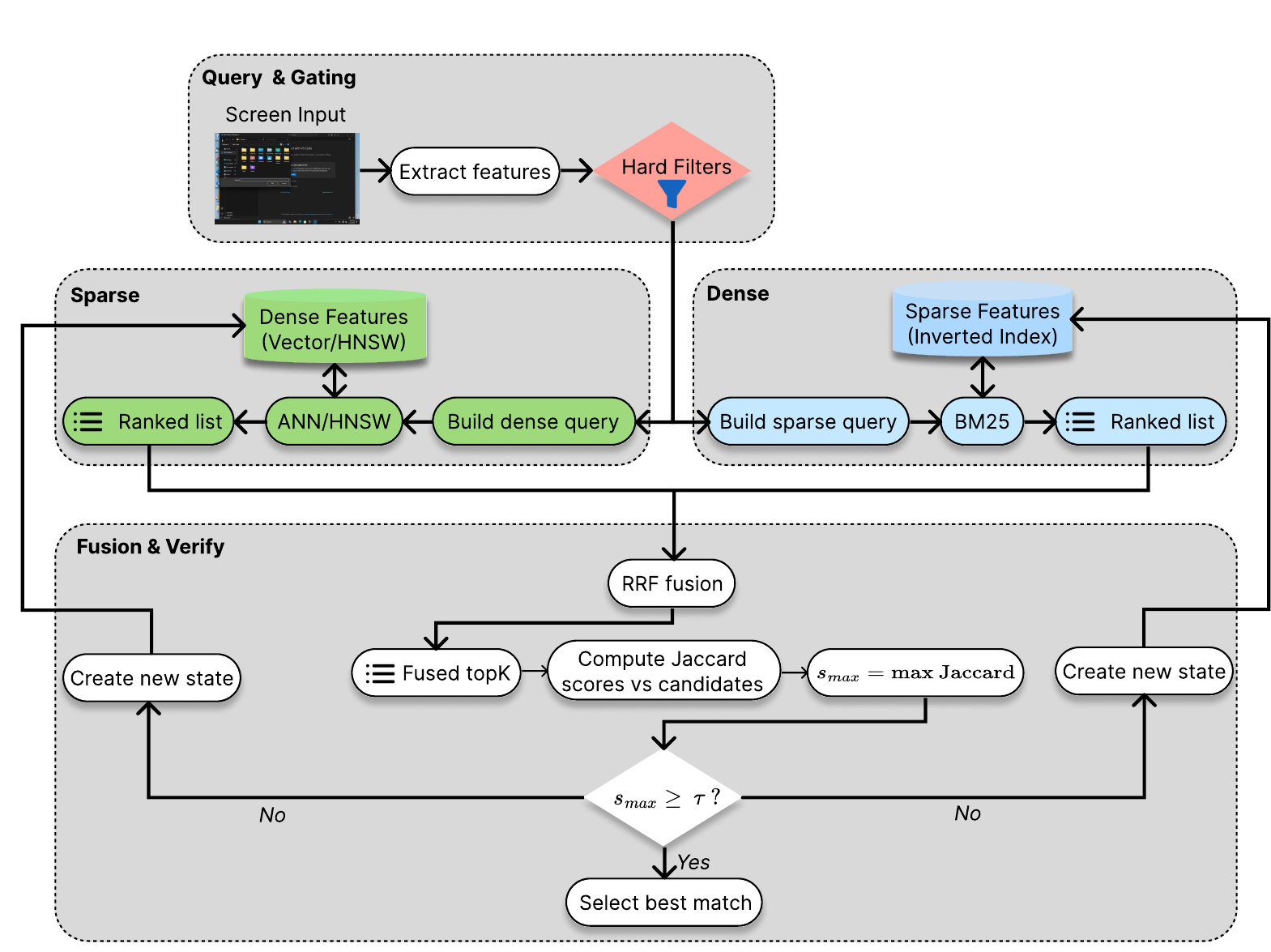}
    \caption{Hybrid screen retrieval. Metadata filters define eligibility; sparse and dense ranks generate candidates; bounded structural overlap makes the final merge decision.}
    \label{fig:screen_retrieval}
\end{figure}

Clicking Save can be represented as
\begin{center}
\small\nolinkurl{focus_click@1::r12_c7|button|save|notepad::none},
\end{center}
whereas typing ``hello'' is
\begin{center}
\small\nolinkurl{focus_type@1::r15_c15|edit|text_editor|notepad::{"text":"hello"}}.
\end{center}
Changing ``hello'' to ``world'' changes the signature and therefore its edge and ambiguity statistics.

\subsection{Reactive LLM baselines}

Reactive runs use GPT-5.4 nano or mini at temperature 0, provider-default top-$p$, and a 140-token output maximum, with no heuristic fallback. The prompt requires one provided candidate and forbids invented IDs. \texttt{ctx=Yes} includes the last eight executed actions; \texttt{ctx=No} omits action history. Both variants receive the instruction, active-window metadata, counters, and up to 60 candidates. They return a JSON object containing an action, candidate ID, optional text, and a short reason; the executor validates it before use.

\section{Original Notepad diagnostic}
\label{app:notepad_diagnostic}

The original policy diagnostic uses three verified Notepad replay starts and a fixed 50-action budget. It is retained as a transparent slice-level comparison of reactive proposal quality and graph-based search, but its small scale is not used as cross-application evidence.

\subsection{Protocol and curve-derived metrics}

Each episode begins from a verified replayable state reached from a clean snapshot by replaying a deterministic prefix. A state enters the slice only if the stored prefix reliably returns to the intended deduplicated state. We restrict the slice to eligible Notepad states with generation-mode provenance, at least three occurrences, and a completed replay anchor at step 5 or later, then deterministically select three starts.

We compare four reactive baselines that act directly from the current screen with the uncertainty-guided PUCT graph-bandit using the default uniform prior. Every episode uses steps 0--49, begins from the same replayed state, and receives the same public information. Runs are not initialized with states, edges, or search statistics discovered in earlier episodes. Keeping the PUCT prior uniform on this diagnostic separates graph-based selection from the deterministic heuristic and learned proposals evaluated elsewhere.

For a replay-start set $\mathcal{B}$, let $S_t^{(b)}$ be the raw observation at step $t$ in episode $b$ and let $s_t^{(b)}:=\pi(S_t^{(b)})$ be its deduplicated state under the retrieval and deduplication map $\pi$. We define
\begin{equation}
M_V(t):=\frac{1}{|\mathcal{B}|}\sum_{b\in\mathcal{B}}
\left|\{s_\tau^{(b)}:0\leq\tau\leq t\}\right|,
\qquad
\Delta u_t:=\frac{1}{|\mathcal{B}|}\sum_{b\in\mathcal{B}}
\left(u(s_t^{(b)})-u(s_0^{(b)})\right).
\label{eq:notepad_curve_metrics}
\end{equation}
We report $M_V(49)$ and $\Delta u_{49}$ together with the discrete summaries $\sum_{t=0}^{49}M_V(t)$ and $\sum_{t=0}^{49}\Delta u_t$. Higher frontier values are better; lower ambiguity values indicate stronger movement toward lower-dispersion states.

\begin{table}[H]
\centering
\small
\setlength{\tabcolsep}{5pt}
\caption{Exact frontier and ambiguity summaries derived from the fixed three-start Notepad replay-start slice.}
\label{tab:frontier_discovery_replay_start}
\resizebox{\linewidth}{!}{%
\begin{tabular}{lrrrr}
\toprule
Method & $M_V(49)\uparrow$ & Frontier AUC$\uparrow$ & $\Delta u_{49}\downarrow$ & Ambiguity AUC$\downarrow$\\
\midrule
Reactive GPT-5.4 nano, no ctx & 10.00 & 340.33 & $-0.00$ & 5.37\\
Reactive GPT-5.4 nano, ctx & 10.00 & 362.00 & 0.03 & 5.41\\
Reactive GPT-5.4 mini, no ctx & \textbf{13.67} & 362.33 & $-0.09$ & 0.55\\
Reactive GPT-5.4 mini, ctx & 10.00 & \textbf{411.00} & $-0.18$ & $-1.23$\\
PUCT graph-bandit & 10.00 & 253.33 & 0.05 & 1.08\\
\bottomrule
\end{tabular}%
}
\end{table}

\definecolor{curve_reactive_nano_a}{HTML}{1E88E5}
\definecolor{curve_reactive_nano_b}{HTML}{00ACC1}
\definecolor{curve_reactive_mini_a}{HTML}{7E57C2}
\definecolor{curve_reactive_mini_b}{HTML}{5E35B1}
\definecolor{curve_search_no_prior}{HTML}{43A047}

\begin{figure*}[t]
\centering
\begin{minipage}[t]{0.485\textwidth}
\centering
\begin{tikzpicture}
\begin{axis}[
  width=0.96\linewidth,
  height=0.74\linewidth,
  xlabel={Step},
  ylabel={$M_V(t)$},
  xmin=0, xmax=49,
  ymin=0, ymax=14,
  grid=both,
  grid style={line width=0.1pt,draw=gray!20},
  major grid style={line width=0.2pt,draw=gray!35},
  tick align=outside,
  legend style={
    font=\scriptsize,
    draw=none,
    fill=none,
    at={(0.5,1.02)},
    anchor=south,
    legend columns=2,
    /tikz/every even column/.append style={column sep=0.35em}
  },
  legend cell align=left
]
\addplot+[no marks,very thick,color=curve_reactive_nano_a] coordinates {(0,0.0000) (1,0.6667) (2,1.6667) (3,2.3333) (4,2.3333) (5,2.6667) (6,3.0000) (7,3.6667) (8,4.0000) (9,4.3333) (10,4.3333) (11,5.0000) (12,5.3333) (13,5.3333) (14,5.6667) (15,5.6667) (16,6.0000) (17,6.0000) (18,6.0000) (19,6.6667) (20,6.6667) (21,7.0000) (22,7.0000) (23,7.0000) (24,7.6667) (25,7.6667) (26,7.6667) (27,7.6667) (28,8.0000) (29,8.0000) (30,8.6667) (31,9.0000) (32,9.0000) (33,9.0000) (34,9.0000) (35,9.0000) (36,9.0000) (37,9.0000) (38,9.0000) (39,9.0000) (40,9.0000) (41,9.3333) (42,9.3333) (43,9.3333) (44,9.3333) (45,9.6667) (46,9.6667) (47,10.0000) (48,10.0000) (49,10.0000)};
\addlegendentry{5.4n | no ctx}
\addplot+[no marks,very thick,color=curve_reactive_nano_b] coordinates {(0,0.0000) (1,0.6667) (2,1.3333) (3,2.3333) (4,3.0000) (5,4.0000) (6,5.0000) (7,5.3333) (8,5.3333) (9,5.3333) (10,5.6667) (11,6.0000) (12,6.3333) (13,6.6667) (14,6.6667) (15,6.6667) (16,6.6667) (17,6.6667) (18,6.6667) (19,7.3333) (20,7.6667) (21,8.0000) (22,8.0000) (23,8.3333) (24,8.3333) (25,8.3333) (26,8.3333) (27,8.3333) (28,8.3333) (29,8.3333) (30,8.3333) (31,8.3333) (32,8.3333) (33,8.3333) (34,8.3333) (35,8.3333) (36,8.3333) (37,8.6667) (38,9.0000) (39,9.3333) (40,9.3333) (41,9.3333) (42,9.6667) (43,9.6667) (44,9.6667) (45,9.6667) (46,9.6667) (47,10.0000) (48,10.0000) (49,10.0000)};
\addlegendentry{5.4n | ctx}
\addplot+[no marks,very thick,color=curve_reactive_mini_a] coordinates {(0,0.0000) (1,0.6667) (2,1.3333) (3,1.6667) (4,2.3333) (5,3.0000) (6,3.6667) (7,4.0000) (8,4.6667) (9,5.3333) (10,5.6667) (11,5.6667) (12,5.6667) (13,6.0000) (14,6.3333) (15,6.3333) (16,6.3333) (17,6.3333) (18,6.3333) (19,6.3333) (20,6.6667) (21,6.6667) (22,6.6667) (23,6.6667) (24,6.6667) (25,6.6667) (26,7.0000) (27,7.3333) (28,7.3333) (29,7.3333) (30,7.3333) (31,7.6667) (32,8.0000) (33,8.3333) (34,8.6667) (35,8.6667) (36,9.0000) (37,9.0000) (38,9.3333) (39,9.6667) (40,9.6667) (41,10.3333) (42,10.6667) (43,11.3333) (44,12.3333) (45,12.6667) (46,13.0000) (47,13.0000) (48,13.3333) (49,13.6667)};
\addlegendentry{5.4m | no ctx}
\addplot+[no marks,very thick,color=curve_reactive_mini_b] coordinates {(0,0.0000) (1,0.6667) (2,1.6667) (3,2.3333) (4,3.0000) (5,3.6667) (6,4.3333) (7,5.0000) (8,5.6667) (9,6.3333) (10,6.6667) (11,7.3333) (12,7.3333) (13,7.6667) (14,8.0000) (15,8.3333) (16,8.3333) (17,8.6667) (18,9.0000) (19,9.3333) (20,9.3333) (21,9.6667) (22,9.6667) (23,9.6667) (24,9.6667) (25,9.6667) (26,10.0000) (27,10.0000) (28,10.0000) (29,10.0000) (30,10.0000) (31,10.0000) (32,10.0000) (33,10.0000) (34,10.0000) (35,10.0000) (36,10.0000) (37,10.0000) (38,10.0000) (39,10.0000) (40,10.0000) (41,10.0000) (42,10.0000) (43,10.0000) (44,10.0000) (45,10.0000) (46,10.0000) (47,10.0000) (48,10.0000) (49,10.0000)};
\addlegendentry{5.4m | ctx}
\addplot+[no marks,very thick,color=curve_search_no_prior] coordinates {(0,0.0000) (1,0.6667) (2,0.6667) (3,1.0000) (4,1.3333) (5,1.6667) (6,2.0000) (7,2.3333) (8,2.3333) (9,2.3333) (10,2.3333) (11,3.0000) (12,3.0000) (13,3.0000) (14,3.6667) (15,3.6667) (16,3.6667) (17,4.0000) (18,4.3333) (19,4.3333) (20,4.3333) (21,4.6667) (22,5.0000) (23,5.0000) (24,5.0000) (25,5.0000) (26,5.0000) (27,5.3333) (28,5.3333) (29,5.3333) (30,5.6667) (31,6.0000) (32,6.0000) (33,6.0000) (34,6.3333) (35,6.3333) (36,6.6667) (37,7.0000) (38,7.0000) (39,7.3333) (40,7.6667) (41,8.0000) (42,8.3333) (43,8.6667) (44,9.0000) (45,9.0000) (46,9.6667) (47,9.6667) (48,9.6667) (49,10.0000)};
\addlegendentry{PUCT graph-bandit}
\end{axis}
\end{tikzpicture}
\captionof{figure}{Frontier-state trajectories on the fixed three-start Notepad replay-start slice. Search continues to discover frontier states late in the budget, even after some reactive policies plateau.}
\label{fig:frontier_state_curve_compact}
\end{minipage}\hfill
\begin{minipage}[t]{0.485\textwidth}
\centering
\begin{tikzpicture}
\begin{axis}[
  width=0.96\linewidth,
  height=0.74\linewidth,
  xlabel={Step},
  ylabel={$\Delta u_t$},
  xmin=0, xmax=49,
  ymin=-0.35, ymax=0.46,
  grid=both,
  grid style={line width=0.1pt,draw=gray!20},
  major grid style={line width=0.2pt,draw=gray!35},
  tick align=outside,
  legend style={
    font=\scriptsize,
    draw=none,
    fill=none,
    at={(0.5,1.02)},
    anchor=south,
    legend columns=2,
    /tikz/every even column/.append style={column sep=0.35em}
  },
  legend cell align=left
]
\addplot+[no marks,very thick,color=curve_reactive_nano_a] coordinates {(0,0.0000) (1,0.3221) (2,0.2860) (3,0.2611) (4,0.2429) (5,0.2245) (6,0.2295) (7,0.2079) (8,0.2017) (9,0.2088) (10,0.2002) (11,0.1982) (12,0.1814) (13,0.1698) (14,0.1624) (15,0.1585) (16,0.1459) (17,0.1425) (18,0.1196) (19,0.1120) (20,0.1095) (21,0.1116) (22,0.1154) (23,0.1012) (24,0.0899) (25,0.0945) (26,0.0890) (27,0.0886) (28,0.0807) (29,0.0811) (30,0.0630) (31,0.0575) (32,0.0546) (33,0.0503) (34,0.0474) (35,0.0509) (36,0.0413) (37,0.0444) (38,0.0387) (39,0.0319) (40,0.0287) (41,0.0204) (42,0.0278) (43,0.0210) (44,0.0199) (45,0.0091) (46,0.0080) (47,0.0098) (48,0.0058) (49,-0.0006)};
\addlegendentry{5.4n | no ctx}
\addplot+[no marks,very thick,color=curve_reactive_nano_b] coordinates {(0,0.0000) (1,0.3218) (2,0.2745) (3,0.2494) (4,0.2288) (5,0.2655) (6,0.2416) (7,0.2224) (8,0.2047) (9,0.1822) (10,0.1810) (11,0.1715) (12,0.1776) (13,0.1619) (14,0.1455) (15,0.1367) (16,0.1311) (17,0.1256) (18,0.1201) (19,0.1168) (20,0.1378) (21,0.1090) (22,0.0994) (23,0.0848) (24,0.0832) (25,0.0751) (26,0.0814) (27,0.0754) (28,0.0615) (29,0.0594) (30,0.0559) (31,0.0495) (32,0.0489) (33,0.0497) (34,0.0435) (35,0.0427) (36,0.0415) (37,0.0342) (38,0.0698) (39,0.0864) (40,0.0394) (41,0.0361) (42,0.0366) (43,0.0367) (44,0.0337) (45,0.0307) (46,0.0408) (47,0.0353) (48,0.0334) (49,0.0345)};
\addlegendentry{5.4n | ctx}
\addplot+[no marks,very thick,color=curve_reactive_mini_a] coordinates {(0,0.0000) (1,0.3218) (2,0.2929) (3,0.2594) (4,0.2389) (5,0.2198) (6,0.1889) (7,0.1707) (8,0.1672) (9,0.1517) (10,0.1333) (11,0.1122) (12,0.0952) (13,0.0898) (14,0.0689) (15,0.0622) (16,0.0273) (17,0.0196) (18,0.0064) (19,0.0045) (20,-0.0099) (21,-0.0049) (22,-0.0087) (23,-0.0104) (24,-0.0141) (25,-0.0158) (26,-0.0287) (27,-0.0331) (28,-0.0607) (29,-0.0621) (30,-0.0636) (31,-0.0653) (32,-0.0753) (33,-0.0792) (34,-0.0920) (35,-0.0993) (36,-0.0972) (37,-0.1131) (38,-0.1224) (39,-0.1336) (40,-0.1417) (41,-0.1280) (42,-0.1089) (43,-0.0719) (44,-0.0244) (45,-0.0803) (46,-0.0641) (47,-0.0938) (48,-0.0881) (49,-0.0886)};
\addlegendentry{5.4m | no ctx}
\addplot+[no marks,very thick,color=curve_reactive_mini_b] coordinates {(0,0.0000) (1,0.3218) (2,0.2030) (3,0.2459) (4,0.2203) (5,0.2036) (6,0.2078) (7,0.1977) (8,0.1799) (9,0.1685) (10,0.1452) (11,0.1072) (12,0.0884) (13,0.0776) (14,0.0634) (15,0.0367) (16,0.0192) (17,0.0241) (18,0.0073) (19,-0.0080) (20,-0.0237) (21,-0.0180) (22,-0.0410) (23,-0.0486) (24,-0.0605) (25,-0.0904) (26,-0.1094) (27,-0.1088) (28,-0.1072) (29,-0.1046) (30,-0.1132) (31,-0.1205) (32,-0.1158) (33,-0.1129) (34,-0.1101) (35,-0.1156) (36,-0.1500) (37,-0.1562) (38,-0.1713) (39,-0.1667) (40,-0.1681) (41,-0.1672) (42,-0.1663) (43,-0.1621) (44,-0.1606) (45,-0.1600) (46,-0.1712) (47,-0.1783) (48,-0.1786) (49,-0.1787)};
\addlegendentry{5.4m | ctx}
\addplot+[no marks,very thick,color=curve_search_no_prior] coordinates {(0,0.0000) (1,0.0303) (2,-0.0246) (3,0.0664) (4,0.0276) (5,0.0510) (6,0.0404) (7,0.0524) (8,0.0307) (9,-0.0384) (10,-0.0412) (11,0.0162) (12,0.0082) (13,-0.0487) (14,0.0080) (15,-0.0531) (16,0.0050) (17,0.0054) (18,0.0072) (19,0.0279) (20,0.0241) (21,0.0669) (22,0.0749) (23,0.0273) (24,0.0784) (25,0.0147) (26,-0.0002) (27,0.0590) (28,0.0404) (29,-0.0067) (30,0.0373) (31,0.0495) (32,0.0373) (33,0.0277) (34,0.0315) (35,0.0289) (36,0.0415) (37,0.0058) (38,0.0290) (39,0.0348) (40,0.0327) (41,0.0308) (42,0.0290) (43,0.0193) (44,-0.0177) (45,0.0087) (46,0.0151) (47,0.0071) (48,0.0278) (49,0.0524)};
\addlegendentry{PUCT graph-bandit}
\end{axis}
\end{tikzpicture}
\captionof{figure}{Ambiguity change on the same slice. Lower $\Delta u_t$ means stronger disambiguation, not greater frontier growth.}
\label{fig:notepad_ambiguity_reduction_curve}
\end{minipage}
\end{figure*}

\paragraph{Interpretation.}
\Cref{tab:frontier_discovery_replay_start,fig:frontier_state_curve_compact,fig:notepad_ambiguity_reduction_curve} show substantial variation across policies on this slice. The reactive GPT-5.4 mini variants attain the largest frontier totals; the context-enabled mini policy has the best frontier AUC and the largest final ambiguity reduction. The nano variants are weaker on both frontier metrics, while the uniform-prior PUCT graph-bandit is more conservative here but continues discovering states late in the budget. Frontier growth and ambiguity reduction should therefore be read as separate diagnostics, and this three-start slice is evidence about behavioral differences rather than uniform method dominance.

\subsection{Learned-proposal control}

The same diagnostic also supports a proposal-prior control. Replacing the uniform PUCT proposal with the learned diffusion proposal raises both state and edge discovery while reducing revisit and loop/stall rates.

\begin{table}[H]
\centering
\small
\setlength{\tabcolsep}{4pt}
\caption{Additional learned-proposal control on the three-start Notepad diagnostic.}
\label{tab:notepad_prior}
\resizebox{\linewidth}{!}{%
\begin{tabular}{llrrrrrr}
\toprule
Selector & Prior & States & State AUC & Edges & Edge AUC & Revisit & Loop/stall\\
\midrule
PUCT & Uniform & 10.00 & 253.33 & 46.67 & 1,199.00 & 0.796 & 0.440\\
PUCT & Learned diffusion & 14.33 & 406.67 & 49.00 & 1,225.00 & 0.708 & 0.253\\
\bottomrule
\end{tabular}%
}
\end{table}

\section{Ambiguity audits}
\label{app:ambiguity_audits}

For 1,588 repeated state--action groups, we decompose pooled dispersion into within-recorded-context and between-recorded-context contributions.

\begin{table}[H]
\centering
\small
\caption{Context decomposition of repeated matched-action dispersion.}
\label{tab:context_decomposition}
\begin{tabular}{lrrrr}
\toprule
Application & Groups & Pooled & Within context & Between context\\
\midrule
Notepad & 715 & 0.158 & 0.005 & 0.152\\
File Explorer & 611 & 0.530 & 0.001 & 0.529\\
LibreOffice Calc & 262 & 0.120 & 0.000 & 0.120\\
\bottomrule
\end{tabular}
\end{table}

The decomposition is descriptive. Recorded context may correlate with hidden workflow state, but it is not a complete latent-state annotation; unrecorded asynchronous effects and state-matching errors remain possible.

\section{Hyperparameter sensitivity}
\label{app:sensitivity}

\Cref{tab:full_component_metrics} gives the complete metric set for the component suite condensed in \cref{tab:component_ablation}.

\begin{table}[H]
\centering
\scriptsize
\caption{Full three-application macro-average component metrics.}
\label{tab:full_component_metrics}
\resizebox{\linewidth}{!}{%
\begin{tabular}{lrrrrrrrr}
\toprule
Configuration & New states & State AUC & New edges & Edge AUC & Net $\Delta u$ & High-$u$ occ. & Revisit & Loop/stall\\
\midrule
Full: novelty + ambiguity & 1.573 & 50.487 & 25.727 & 716.027 & 0.0083 & 0.6568 & 0.9678 & 0.8593\\
Novelty only & 1.427 & 46.347 & 31.153 & 901.533 & -0.0346 & 0.4907 & 0.9705 & 0.8828\\
Ambiguity only & 1.047 & 34.300 & 26.640 & 761.987 & 0.0050 & 0.6818 & 0.9786 & 0.9000\\
Prior only & 2.233 & 71.867 & 36.300 & 982.900 & 0.0218 & 0.7068 & 0.9544 & 0.8153\\
Full + uniform prior & 4.733 & 132.527 & 41.880 & 1,094.720 & 0.0343 & 0.7983 & 0.9028 & 0.6937\\
Full + exact matching & 2.100 & 67.847 & 36.227 & 997.240 & -0.0155 & 0.6038 & 0.9570 & 0.8651\\
\bottomrule
\end{tabular}
}
\end{table}

Each row in \cref{tab:sensitivity} varies one parameter while holding all other settings fixed in the matched three-application suite.

\begin{table}[H]
\centering
\small
\caption{One-factor-at-a-time sensitivity.}
\label{tab:sensitivity}
\begin{tabular}{llll}
\toprule
Parameter & Values & New-state AUC & Diagnostic\\
\midrule
$\tau$ & 0.90 / 0.93 / 0.96 & 54.573 / 50.487 / 33.847 & False merge 0.009 / 0.005 / 0.004\\
$\kappa$ & 5 / 20 / 50 & 46.901 / 50.487 / 47.353 & Revisit 0.9714 / 0.9678 / 0.9688\\
$u_0$ & 0.25 / 0.50 / 0.75 & 31.073 / 50.487 / 65.307 & Revisit 0.9781 / 0.9678 / 0.9554\\
\bottomrule
\end{tabular}
\end{table}

For $\tau=0.90/0.93/0.96$, pair F1 is 0.632/0.622/0.594 and net ambiguity change is 0.0164/0.0083/$-0.0371$. Higher $\tau$ reduces false merges but creates more fragmented identities; the persisted UIA IDs used for pair F1 are a matcher diagnostic, not independent semantic ground truth.

\section{Alternative observation representation}
\label{app:representation_audit}

We re-encode 1,200 stored screenshots---450 Notepad, 450 File Explorer, and 300 LibreOffice Calc---using offline UIA features or spatially localized OCR text. Existing \algacro{} state IDs serve as reference labels. Within each application and representation, 40\% of pairs select a cutoff that is frozen for the remaining 60\%. RapidOCR 1.4.4 with PP-OCRv4 returns non-empty OCR for every screenshot.

\begin{table}[H]
\centering
\small
\setlength{\tabcolsep}{4pt}
\caption{UIA and OCR representations on matched stored screenshots and actions.}
\label{tab:ocr_audit}
\begin{tabular}{lrrrr}
\toprule
Representation & Pair F1 & Recall@10 & False merge & Successor consistency\\
\midrule
Offline UIA & 0.740 & 0.381 & 0.010 & 0.892\\
OCR only & 0.646 & 0.409 & 0.007 & 0.958\\
\bottomrule
\end{tabular}
\end{table}

\section{Downstream training details}
\label{app:downstream_details}

Task descriptions are synthesized with \texttt{gpt-4o-mini}; the model is used for label generation rather than as the PUCT selector. The reactive GPT-5.4 baselines described in \cref{app:reproducibility} are evaluation comparators.

\begin{table}[H]
\centering
\small
\caption{Downstream training-data and evaluation properties.}
\begin{tabular}{lr}
\toprule
Property & Value\\
\midrule
Three-step trajectory windows & 22,028\\
Unique task descriptions & 8,118\\
Semantically unique descriptions & 75.4\%\\
Windows containing a state change & 93.99\%\\
Frozen evaluation prompts & 1,000\\
\bottomrule
\end{tabular}
\end{table}

Each action stores operation, target role/text, position, and optional argument. The model predicts the next typed action from the task and current screen. Counterfactual pairwise supervision pairs examples with the same task but different screens and encourages different actions when the GUI state requires them. Step Action Match is exact agreement with the recorded action. Screen Sensitivity is the accuracy decrease after shuffling screen context; State-Change Sensitivity applies the same test to windows containing a GUI state change. Offline three-step success and post-change recovery use recorded next screens after each prediction and are therefore teacher-forced proxies.

\subsection{Software assets and licenses}

RapidOCR and PaddleOCR are used under the Apache License 2.0. The Visual Studio Code source repository is MIT-licensed, and LibreOffice is distributed under MPL-2.0/LGPLv3+ terms. Commercial Microsoft applications and OpenAI API models are accessed under the authors' applicable organizational licenses and service terms. We do not redistribute third-party application code or model weights.

\section{Additional corpus diagnostics}

\subsection{Screenshot post-processing}
\label{appendix:screenshot_post_processing}
\begin{figure}[h]
\centering
\includegraphics[width=\linewidth,height=0.18\textheight,keepaspectratio]{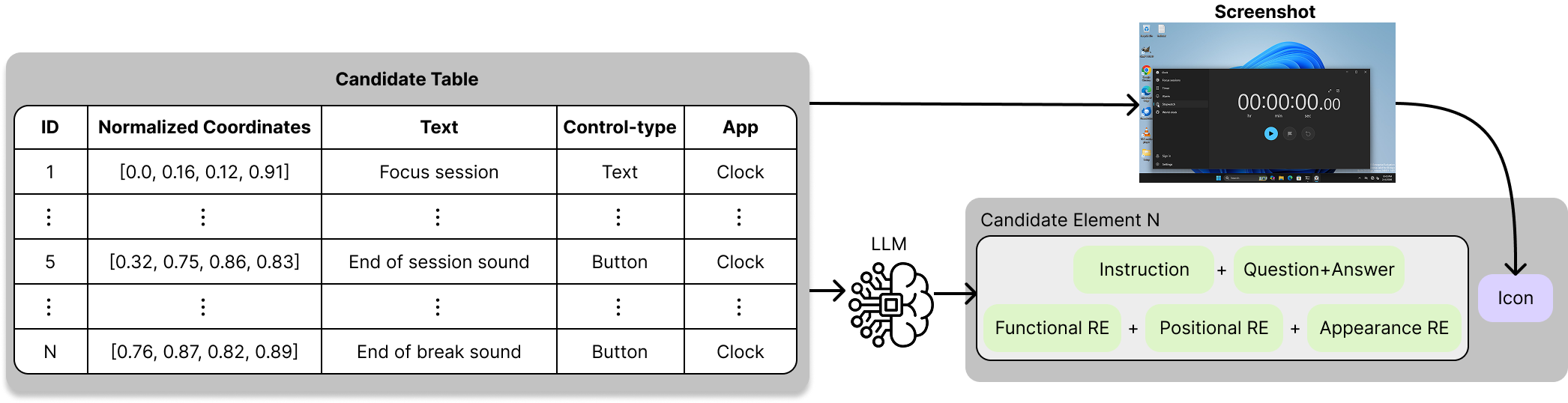}
\caption{Overview of screenshot post-processing pipeline.}
\label{fig:screenshot_post_processing}
\end{figure}

Each exploration step logs a raw screenshot alongside its structured UIA-derived candidate table, enabling fully offline annotation over the collected corpus. As shown in Figure~\ref{fig:screenshot_post_processing}, we then run a lightweight screenshot post-processing pipeline that takes the candidate table as input and emits a compact set of textual labels attached to the screenshot. These labels include a natural-language instruction, a question--answer pair, and short rationales that explain the screen in functional, positional, and appearance terms. We additionally map screenshot regions to icon instances in the candidate table to generate labels that are explicitly grounded to a particular icon.

Importantly, this pipeline does \emph{not} require a vision-language model: the only model dependency is an LLM used to generate text labels from the structured UI metadata. Because post-processing is decoupled from exploration and operates over normalized, schema-consistent artifacts, new labelers and additional post-processing procedures can be implemented as standalone passes and replayed over existing data without re-running VM rollouts. The exact prompt templates are provided in Appendix~\ref{appendix:processing_prompts}.

\subsection{Screenshot processing prompts}
\label{appendix:processing_prompts}

\begingroup
\renewcommand{\arraystretch}{1.15}
\small

\captionsetup{type=table}
\captionof{table}{Prompt templates for icon/element labeling. Each prompt is conditioned on the UIA-derived candidate table. The model receives one or more candidate rows describing the target icon/element.}
\label{tab:icon_label_prompt_templates}

\begin{tabularx}{\textwidth}{lX}
\toprule
\textbf{Component} & \textbf{Prompt Template} \\
\midrule

Caption &
\begin{minipage}[t]{\linewidth}\ttfamily
ID | Type | Text content or description | Normalized location [x1, y1, x2, y2]\\
\textless CANDIDATE\_ITEM\_1\textgreater \ \ \textit{(e.g., \textless ID\textgreater\ \textless TYPE\textgreater\ \textless TEXT\_OR\_DESC\textgreater\ [\textless x1\textgreater,\textless y1\textgreater,\textless x2\textgreater,\textless y2\textgreater])}\\
The attached text data describes a specific icon/element present in a computer screen.
Provide a concise 1--2 sentence caption describing this icon/element with respect to the entire computer screen.
\end{minipage}
\\
\midrule

Instruction &
\begin{minipage}[t]{\linewidth}\ttfamily
ID | Type | Text content or description | Normalized location [x1, y1, x2, y2]\\
\textless CANDIDATE\_ITEM\_1\textgreater\\
The attached text data describes a specific icon/element present in the computer screen.
Provide a concise 1--2 sentence instruction describing how to use this icon/element with respect to the entire computer screen.
\end{minipage}
\\
\midrule

Question+Answer &
\begin{minipage}[t]{\linewidth}\ttfamily
ID | Type | Text content or description | Normalized location [x1, y1, x2, y2]\\
\textless CANDIDATE\_ITEM\_1\textgreater\\
The attached text data describes a specific icon/element present in the computer screen.
Provide a concise natural language question a user might ask about using this icon/element.
Also, provide the expected answer to the question.
\end{minipage}
\\
\midrule

All REs &
\begin{minipage}[t]{\linewidth}\ttfamily
ID | Type | Text content or description | Normalized location [x1, y1, x2, y2]\\
\textless CANDIDATE\_ITEM\_1\textgreater\\
I will provide data above about a specific icon/element to help you understand it.
Your task is to classify the target area and generate several references regarding the target area on the original computer screen image.\\[2pt]
Specific task requirements are as follows: You will analyze and output a dictionary in JSON file format.
The key-value pairs included are as follows:\\
- functional\_reference: A reference about the target area, involving the function of the target area.\\
- positional\_reference: A reference about the target area by describing the position of the target area, such as layout and nearby elements.\\
- appearance\_reference: A reference about the target area by describing the appearance of the target area.\\[2pt]
\textbf{Ensure that anyone can uniquely identify this area in the screenshot through any one of the references.}\\
\textbf{Ensure that the reference is complete and independent.}\\[2pt]
Your output should follow this format strictly:\\
\{``Output'': \{``functional\_reference'': ``\textless ...\textgreater'', ``positional\_reference'': ``\textless ...\textgreater'', ``appearance\_reference'': ``\textless ...\textgreater''\}\}
\end{minipage}
\\

\bottomrule
\end{tabularx}
\endgroup

\subsection{Action tokenization}
\label{appendix:action_tokenization}

\subsection{Token normalization and rule-based semantic/value classes}
\paragraph{Text normalization.}
All role/semantic strings are normalized by lowercasing, collapsing whitespace, and truncating to 120 characters.

\paragraph{Value-class mapping for \texttt{type}.}
Given a normalized typed string $u$, we map:
\begin{itemize}
  \item \texttt{val\_email} if $u$ contains ``@'' or the substring ``email'';
  \item \texttt{val\_password} if $u$ contains ``pass'' or ``pwd'';
  \item \texttt{val\_search} if $u$ contains ``search'', ``find'', or ``query'';
  \item otherwise \texttt{val\_generic\_text}.
\end{itemize}

\paragraph{Position binning from \texttt{location}.}
If \texttt{location} matches \texttt{r<i>\_c<j>} with $i,j\in[0,29]$, we bin:
\[
\texttt{row\_bin}(i)=\texttt{top}\ (i<10),\ \texttt{mid}\ (10\le i<20),\ \texttt{bot}\ (i\ge 20),
\]
\[
\texttt{col\_bin}(j)=\texttt{left}\ (j<10),\ \texttt{center}\ (10\le j<20),\ \texttt{right}\ (j\ge 20),
\]
and emit \texttt{row\_bin\_col\_bin}; otherwise \texttt{unknown}.
\paragraph{Semantic class mapping for $c_t$.}
We derive $c_t$ from the \texttt{semantic} string associated with the action target using a small set of special cases,
regex detectors, and a fixed substring-based taxonomy (default \texttt{generic}).

\begin{itemize}
  \item Special cases: \texttt{settings} $\rightarrow$ \texttt{settings\_root}; glyph-only strings $\rightarrow$ \texttt{icon}.
  \item Network patterns: IPv4 / MAC / IPv6-like strings $\rightarrow$ \texttt{network\_value}.
  \item Otherwise, we assign the first matching class in Table~\ref{tab:semantic_class_rules} via substring rules (fallback \texttt{generic}).
\end{itemize}

\begin{table}[t]
\centering
\small
\setlength{\tabcolsep}{6pt}
\begin{tabularx}{\linewidth}{lX}
\toprule
\textbf{Class $c_t$} & \textbf{Matched text/semantic items (substring triggers)} \\
\midrule
\texttt{search} & \texttt{search box}, \texttt{find a setting}, \texttt{show all results}, \texttt{expand search box}, \texttt{search}, \texttt{find}, \texttt{query} \\
\texttt{window\_control} & \texttt{minimize settings}, \texttt{maximize settings}, \texttt{restore settings}, \texttt{close settings}, \texttt{minimize}, \texttt{maximize}, \texttt{restore} \\
\texttt{navigation} & \texttt{open navigation}, \texttt{back}, \texttt{home} \\
\texttt{settings\_recommendation} & \texttt{recommended settings}, \texttt{commonly used settings}, \texttt{optimize your settings} \\
\texttt{settings\_network} & \texttt{network \& internet}, \texttt{ethernet}, \texttt{bluetooth}, \texttt{dns}, \texttt{gateway}, \texttt{ipv4}, \texttt{ipv6}, \texttt{adapter}, \texttt{internet} \\
\texttt{settings\_update} & \texttt{windows update}, \texttt{update history}, \texttt{up to date} \\
\texttt{settings\_personalization} & \texttt{personalization}, \texttt{background}, \texttt{theme}, \texttt{color mode}, \texttt{dark}, \texttt{lock screen}, \texttt{pictures} \\
\texttt{settings\_accessibility} & \texttt{accessibility}, \texttt{text size}, \texttt{text cursor}, \texttt{cursor thickness}, \texttt{make text size bigger} \\
\texttt{settings\_accounts} & \texttt{accounts}, \texttt{sign-in options}, \texttt{windows hello}, \texttt{family}, \texttt{other users}, \texttt{local account}, \texttt{profile picture} \\
\texttt{settings\_apps} & \texttt{installed apps}, \texttt{apps} \\
\texttt{settings\_time\_language} & \texttt{time \& language} \\
\texttt{settings\_privacy\_security} & \texttt{privacy \& security} \\
\texttt{settings\_gaming} & \texttt{gaming} \\
\texttt{settings\_printers} & \texttt{printers \& scanners} \\
\texttt{settings\_system} & \texttt{system} \\
\texttt{settings\_device\_info} & \texttt{device info}, \texttt{specifications}, \texttt{manufacturer}, \texttt{driver version}, \texttt{physical address} \\
\texttt{status\_offline} & \texttt{no internet}, \texttt{not connected}, \texttt{no\_internet} \\
\texttt{action\_delete} & \texttt{delete} \\
\texttt{action\_rename} & \texttt{rename} \\
\texttt{action\_browse} & \texttt{browse} \\
\texttt{email} & \texttt{email}, \texttt{@} \\
\texttt{password} & \texttt{password}, \texttt{pwd} \\
\texttt{continue} & \texttt{continue}, \texttt{submit}, \texttt{next} \\
\texttt{signin} & \texttt{sign in}, \texttt{log in}, \texttt{login}, \texttt{sign-in} \\
\texttt{cancel} & \texttt{cancel} \\
\texttt{confirm} & \texttt{confirm}, \texttt{ok} \\
\texttt{save} & \texttt{save} \\
\texttt{open} & \texttt{open} \\
\texttt{close} & \texttt{close} \\
\bottomrule
\end{tabularx}
\caption{Rule-based mapping from target semantic text to $c_t$ (first-match wins; fallback \texttt{generic}).}
\label{tab:semantic_class_rules}
\end{table}

\subsection{Action-token distribution}
\label{app:action_token_marginals}
\begin{table}[!ht]
\centering
\caption{Marginal frequencies for each action-slot token in the training corpus.}
\label{tab:action_token_marginals}
\renewcommand{\arraystretch}{1.05}
\setlength{\tabcolsep}{6pt}
\begin{tabular}{ll}
\hline
\rowcolor{gray!20}
\textbf{Metric} & \textbf{Value} \\
\hline

\rowcolor{gray!10}\multicolumn{2}{l}{\textbf{\texttt{x.op} values}} \\
\texttt{x.op}=\texttt{click} & 63{,}881 (96.67\%) \\
\texttt{x.op}=\texttt{key} & 2{,}150 (3.25\%) \\
\texttt{x.op}=\texttt{unknown} & 50 (0.08\%) \\
\texttt{x.op}=\texttt{scroll} & 3 (0.00\%) \\
\hline

\rowcolor{gray!10}\multicolumn{2}{l}{\textbf{\texttt{x.delta.r} values}} \\
\texttt{x.delta.r}=\texttt{text} & 45{,}329 (68.59\%) \\
\texttt{x.delta.r}=\texttt{image} & 12{,}592 (19.05\%) \\
\texttt{x.delta.r}=\texttt{button} & 5{,}868 (8.88\%) \\
\texttt{x.delta.r}=\texttt{edit} & 2{,}150 (3.25\%) \\
\texttt{x.delta.r}=\texttt{pane} & 73 (0.11\%) \\
\texttt{x.delta.r}=\texttt{window} & 50 (0.08\%) \\
\texttt{x.delta.r}=\texttt{menuitem} & 19 (0.03\%) \\
\texttt{x.delta.r}=\texttt{list} & 3 (0.00\%) \\
\hline

\rowcolor{gray!10}\multicolumn{2}{l}{\textbf{\texttt{x.delta.c} values}} \\
\texttt{x.delta.c}=\texttt{icon} & 11{,}242 (17.01\%) \\
\texttt{x.delta.c}=\texttt{unlabeled\_image} & 11{,}178 (16.91\%) \\
\texttt{x.delta.c}=\texttt{settings\_network} & 6{,}150 (9.31\%) \\
\texttt{x.delta.c}=\texttt{settings\_accounts} & 5{,}394 (8.16\%) \\
\texttt{x.delta.c}=\texttt{generic} & 4{,}791 (7.25\%) \\
\texttt{x.delta.c}=\texttt{settings\_update} & 4{,}164 (6.30\%) \\
\texttt{x.delta.c}=\texttt{search} & 3{,}443 (5.21\%) \\
\texttt{x.delta.c}=\texttt{settings\_apps} & 3{,}248 (4.91\%) \\
\texttt{x.delta.c}=\texttt{settings\_system} & 2{,}655 (4.02\%) \\
\texttt{x.delta.c}=\texttt{settings\_root} & 2{,}221 (3.36\%) \\
\texttt{x.delta.c}=\texttt{settings\_time\_language} & 2{,}198 (3.33\%) \\
\texttt{x.delta.c}=\texttt{settings\_personalization} & 1{,}986 (3.01\%) \\
\texttt{x.delta.c}=\texttt{navigation} & 1{,}481 (2.24\%) \\
\texttt{x.delta.c}=\texttt{settings\_accessibility} & 1{,}409 (2.13\%) \\
\texttt{x.delta.c}=\texttt{unlabeled\_button} & 1{,}323 (2.00\%) \\
\texttt{x.delta.c}=\texttt{settings\_privacy\_security} & 863 (1.31\%) \\
\texttt{x.delta.c}=\texttt{settings\_gaming} & 570 (0.86\%) \\
\texttt{x.delta.c}=\texttt{settings\_sorting} & 308 (0.47\%) \\
\texttt{x.delta.c}=\texttt{settings\_device\_info} & 221 (0.33\%) \\
\texttt{x.delta.c}=\texttt{cancel} & 219 (0.33\%) \\
\hline

\rowcolor{gray!10}\multicolumn{2}{l}{\textbf{\texttt{x.delta.p} values}} \\
\texttt{x.delta.p}=\texttt{top\_left} & 17{,}177 (25.99\%) \\
\texttt{x.delta.p}=\texttt{mid\_left} & 14{,}176 (21.45\%) \\
\texttt{x.delta.p}=\texttt{top\_center} & 7{,}709 (11.67\%) \\
\texttt{x.delta.p}=\texttt{bot\_left} & 6{,}986 (10.57\%) \\
\texttt{x.delta.p}=\texttt{bot\_right} & 4{,}542 (6.87\%) \\
\texttt{x.delta.p}=\texttt{top\_right} & 4{,}365 (6.61\%) \\
\texttt{x.delta.p}=\texttt{mid\_center} & 3{,}918 (5.93\%) \\
\texttt{x.delta.p}=\texttt{bot\_center} & 3{,}732 (5.65\%) \\
\texttt{x.delta.p}=\texttt{mid\_right} & 3{,}332 (5.04\%) \\
\texttt{x.delta.p}=\texttt{unknown} & 147 (0.22\%) \\
\hline

\rowcolor{gray!10}\multicolumn{2}{l}{\textbf{\texttt{x.eta} values}} \\
\texttt{x.eta}=\texttt{(empty)} & 63{,}931 (96.74\%) \\
\texttt{x.eta}=\texttt{computer.keyboard.write("text")} & 2{,}150 (3.25\%) \\
\texttt{x.eta}=\texttt{down:med} & 3 (0.00\%) \\
\hline
\end{tabular}
\end{table}

\subsection{Confusion analysis}
\label{app:confusion_analysis_tables}

\noindent
\begin{center}
\renewcommand{\arraystretch}{1.05}
\setlength{\tabcolsep}{6pt}
\captionsetup{skip=2pt}
\footnotesize
\begin{tabular}{llccc}
\hline
\rowcolor{gray!20}
\textbf{Slot} & \textbf{Class} & \textbf{Errors} & \textbf{Share} & \textbf{Class err. rate} \\
\hline
$r$ & \texttt{text}   & 41 & 46.6\% & 19.0\% \\
$r$ & \texttt{image}  & 26 & 29.5\% & 50.0\% \\
$r$ & \texttt{button} & 14 & 15.9\% & 56.0\% \\
$r$ & \texttt{edit}   & 7  & 8.0\%  & 100.0\% \\
\hline
$c$ & \texttt{icon}            & 26 & 22.2\% & 41.9\% \\
$c$ & \texttt{unlabeled\_image}& 21 & 17.9\% & 44.7\% \\
$c$ & \texttt{settings\_apps}  & 10 & 8.5\%  & 47.6\% \\
$c$ & \texttt{settings\_system}& 10 & 8.5\%  & 55.6\% \\
$c$ & \texttt{generic}         & 9  & 7.7\%  & 52.9\% \\
\hline
$p$ & \texttt{top\_left}   & 25 & 21.2\% & 29.1\% \\
$p$ & \texttt{mid\_left}   & 23 & 19.5\% & 41.8\% \\
$p$ & \texttt{top\_center} & 21 & 17.8\% & 60.0\% \\
$p$ & \texttt{bot\_left}   & 15 & 12.7\% & 38.5\% \\
$p$ & \texttt{bot\_right}  & 10 & 8.5\%  & 37.0\% \\
\hline
\end{tabular}
\captionof{table}{Dominant error-driving classes by slot (shares are within-slot).}
\label{tab:slot_error_classes}
\vspace{-0.6em}
\end{center}

\noindent
\begin{center}
\renewcommand{\arraystretch}{1.05}
\setlength{\tabcolsep}{6pt}
\captionsetup{skip=2pt}
\footnotesize
\begin{tabular}{llcc}
\hline
\rowcolor{gray!20}
\textbf{Slot} & \textbf{Confusion} & \textbf{Count} & \textbf{Share of slot errors} \\
\hline
$r$ & \texttt{text$\rightarrow$image}   & 31 & 35.2\% \\
$r$ & \texttt{image$\rightarrow$text}   & 26 & 29.5\% \\
$r$ & \texttt{button$\rightarrow$text}  & 14 & 15.9\% \\
$r$ & \texttt{text$\rightarrow$button}  & 10 & 11.4\% \\
$r$ & \texttt{edit$\rightarrow$text}    & 5  & 5.7\% \\
\hline
$c$ & \texttt{icon$\rightarrow$unlabeled\_image}            & 7 & 6.0\% \\
$c$ & \texttt{unlabeled\_image$\rightarrow$icon}            & 5 & 4.3\% \\
$c$ & \texttt{unlabeled\_image$\rightarrow$settings\_system} & 4 & 3.4\% \\
$c$ & \texttt{settings\_apps$\rightarrow$icon}              & 4 & 3.4\% \\
$c$ & \texttt{settings\_accounts$\rightarrow$search}        & 3 & 2.6\% \\
\hline
$p$ & \texttt{mid\_left$\rightarrow$top\_left}    & 8 & 6.8\% \\
$p$ & \texttt{top\_left$\rightarrow$mid\_left}    & 8 & 6.8\% \\
$p$ & \texttt{top\_left$\rightarrow$top\_center}  & 7 & 5.9\% \\
$p$ & \texttt{bot\_left$\rightarrow$top\_left}    & 6 & 5.1\% \\
$p$ & \texttt{mid\_left$\rightarrow$bot\_center}  & 5 & 4.2\% \\
\hline
\end{tabular}
\captionof{table}{Top within-slot confusions under best-of-$k$ decoding.}
\label{tab:slot_top_confusions}
\vspace{-0.6em}
\end{center}

\end{document}